# Integrated Guidance and Control of a Mother–Child UAV–UGV System for Cooperative Missions

Aashish Sahu[1] | R. Prasanth Kumar[2]

[1]Department of Mechanical & Aerospace Engineering, Indian Institute of Technology Hyderabad, Telangana, India | [2]Department of Artificial Intelligence, Indian Institute of Technology Hyderabad, Telangana, India |

**Correspondence:** R. Prasanth Kumar (rpkumar@ai.iith.ac.in)

## ABSTRACT

Autonomous recovery of a small multirotor onto a hovering multirotor carrier differs fundamentally from recovery onto a ground or shipborne platform because the recovery surface is an actively controlled, thrust-limited aerial vehicle whose motion enters the terminal relative-motion problem. This paper presents an end-to-end, field-validated autonomy framework for a heterogeneous rover–mothership–child system that executes rover supervision, mothership transit, child deployment and sortie, autonomous return, aerial recovery, and synchronized descent. The recovery stack combines jerk-bounded phase reference generation, disturbance-observer-augmented planar tracking, feasibility-aware vertical command generation, a discrete-time barrier-based safety filter for relative vertical geometry, and communication-aware carrier-state prediction. The individual components are established methods; the contribution lies in their coordinated system-level composition for recovery onto a hovering multirotor carrier and in full-scale outdoor validation. The framework is implemented on a PX4–ROS 2 architecture using RTK-enabled GNSS, IMU, and barometric fusion, with a mothership-side 1D lidar used only as an auxiliary near-contact cue. RTK-fixed positioning was maintained throughout the campaign. Across 20 outdoor cooperative mission trials, 17 completed the deployment–sortie–recovery sequence successfully, corresponding to an observed mission success rate of 85%. For the successful recoveries, the mean terminal-alignment time was 6.3 s, the mean planar alignment error at acceptance was 0.18 m, the maximum terminal planar deviation was 0.32 m within a 0.40 m capture radius, and the minimum logged relative vertical separation during coupled descent was 0.41 m. The mothership planar station-keeping RMS error was 0.25 m, approximately 63% of the capture radius. The three unsuccessful trials occurred at different mission stages and are analyzed separately. These results demonstrate practical autonomous aerial recovery within the tested outdoor operating envelope.

## 1 | Introduction

Field robotics increasingly relies on heterogeneous robotic teams in which unmanned ground vehicles (UGVs) and unmanned aerial vehicles (UAVs) cooperate to accomplish demanding missions such as search-and-rescue, infrastructure inspection, environmental monitoring, radiation survey, and reconnaissance. In such teams, UAVs provide rapid deployment, wide-area visibility, and agile sensing, whereas UGVs offer long endurance, greater payload capacity, and a stable base for computation and communications. These complementary capabilities make UAV–UGV cooperation an attractive paradigm for extending mission range, persistence, and operational flexibility in outdoor environments [1, 2, 3, 4, 5, 6, 7]. More broadly, related work in cooperative aerial systems and aerial manipulation also reflects the growing interest in coordinated multi-platform autonomy [8].

A practical realization of this cooperation is the *mothership–child* concept, in which a carrier platform transports a smaller aerial vehicle close to the area of interest, deploys it for a local sortie, and later supports its recovery. In robotics, this concept appears in heterogeneous marsupial systems, deployment–recovery platforms, and aerial transport architectures designed to extend the effective utility of a smaller deployed vehicle [6, 9, 10]. A related viewpoint also appears in the operations-research and logistics literature, where mothership, truck–drone, and hybrid vehicle–drone formulations emphasize range extension, coordinated dispatch, and improved mission efficiency through carrier-supported aerial operations [11, 12, 13, 14, 15]. In addition, launch, docking, and in-flight deployment concepts have been explored in airborne and search-and-rescue settings, further underlining the practical importance of reliable carrier-supported operation [16, 17, 18, 19].

Despite these advantages, the operational benefits of a mothership architecture depend critically on reliable recovery. In field conditions, recovery is often the most demanding phase of the mission because it requires accurate regulation of relative motion, safe proximity operations, and robust execution despite disturbances, intermittent measurements, and communication imperfections. Related localization and coordination challenges have also been recognized in mother–daughter collaborative settings beyond multirotor-only operation. As a result, autonomous recovery should not be treated as a secondary add-on to repeated and extended heterogeneous field operations.

Autonomous landing and docking on moving platforms has therefore received substantial attention in the robotics literature. Prior work has explored vision-based relative localization using fiducial markers or feature-based methods, image-based visual servoing, visual–inertial estimation, and model-based terminal guidance for landing on mobile ground or aerial platforms [20, 9, 21, 22, 23, 24, 25]. Other studies have examined landing under turbulent wind, offshore perception, and dynamic moving-platform recovery, thereby highlighting the practical difficulty of transferring terminal landing methods from controlled experiments to field conditions [26, 27, 10]. These studies have significantly advanced the state of the art, but outdoor deployment still exposes a persistent gap between successful terminal demonstrations and robust field-ready recovery. In practice, docking behavior can become fragile when the controller must operate amid wind disturbances, steady aerodynamic bias, communication delays, or intermittent relative updates.

The subset of this literature in which the recovery platform is itself an airborne rotorcraft is considerably smaller and is directly relevant to the configuration considered in this paper. Dong et al. formulate terminal landing on a midair carrier as an onboard trajectory-optimization problem, using a numerically estimated inter-vehicle aerodynamic-interference model to construct a keep-out region and guided landing cone, and demonstrate centimeter-scale terminal precision in indoor and outdoor experiments [28]. A small quadrotor landing on a substantially larger hexacopter has also been investigated in simulation using radial-basis-function PID trajectory tracking [29]. Complementary ground-test and computational-fluid-dynamics studies of marsupial parent–child rotorcraft identify interference regions above the carrier and associated changes in aerodynamic loading [30]. These studies establish the feasibility of aerial-to-aerial recovery while also showing that the problem cannot be treated simply as a conventional landing with the ground platform replaced by another waypoint.

The comparison with Dong, Li, Cui, Xiang, Li and Tu [28] is particularly useful because it clarifies the scope of the contribution made here. This paper does not claim superior terminal precision. Instead, three configuration-level distinctions are emphasized. First, the child UAV operates using the relative state formed from the two RTK-enabled onboard state estimates and the mothership state exchanged through the inter-vehicle communication architecture, rather than relying on direct visual observation of the carrier. Both RTK solutions remained fixed during the reported recovery campaign; receiver-reported positioning quality, communication timing, and terminal alignment error are therefore treated as distinct experimental quantities rather than being combined into an artificial variance decomposition. Second, the recovery platform is a 14 kg hexacopter whose measured planar station-keeping RMS error is 0.25 m relative to a 0.40 m capture radius, so the child must regulate its terminal motion with respect to an airborne target whose own motion is significant on the scale of the capture region. Third, the experiment addresses the complete supervised deployment–sortie–return–recovery sequence rather than terminal landing in isolation. The contribution is therefore the field-validated realization and evaluation of this integrated aerial-recovery architecture, rather than a claim of improved terminal positioning accuracy over Dong, Li, Cui, Xiang, Li and Tu [28].

From a control perspective, outdoor aerial recovery is simultaneously a disturbance-rejection problem and a constraint-enforcement problem. Persistent wind and aerodynamic bias can induce steady tracking offsets unless they are explicitly estimated and compensated [31, 32, 33]. At the same time, the recovery controller must respect thrust, attitude, and geometric safety limits [34, 35] while operating in a narrow capture region near the docking platform. Control barrier functions provide a principled framework for enforcing safety constraints by modifying nominal commands only as much as needed to preserve forward invariance of a safe set [36, 37]. The individual ingredients required for this task are well established: disturbance-observer and incremental-nonlinear formulations for wind and disturbance rejection [31, 32, 33]; barrier-based safety filtering [36, 37], including discrete-time [38, 39] and high-relative-degree formulations [40]; and delay- and dropout-tolerant coordination for landing on moving platforms [21, 23]. Accordingly, this paper does not claim methodological novelty for these individual components. What is investigated is their coordinated use within a single field-implemented aerial-to-aerial recovery architecture in which disturbance rejection, actuation feasibility, vertical safety, inter-vehicle state exchange, and terminal docking must operate simultaneously. The closest treatments either address terminal aerial landing in isolation [28] or study the corresponding quadrotor-on-hexacopter configuration in simulation [29].

## 1.1 | Why Recovery onto a Hovering Carrier Is Not Recovery onto a Ground Platform

Terminal recovery onto a ground vehicle or ship deck and recovery onto a hovering multirotor share the objective of reducing relative position and velocity at capture, but the physical and control conditions are substantially different. Five characteristics of the experimental platform make this distinction explicit.

First, the recovery surface is actuated and thrust-limited rather than mechanically supported by the ground. A ground platform follows its motion independently of the vehicle attempting to land on it, whereas the mothership simultaneously regulates its own position using finite thrust and attitude authority, while the child UAV converges toward it. The child's terminal reference is therefore associated with the regulated motion of a second aerial vehicle rather than with an effectively rigid surface. The recovery problem is consequently a coupled two-vehicle regulation problem during the final approach and synchronized descent.

Second, deployment and recovery introduce abrupt loading transitions on the carrier. The mothership has an all-up mass of 14.0 kg and the child UAV a mass of 1.8 kg (Table 3), giving a combined pre-deployment mass of 15.8 kg. Deployment therefore removes 11.4% of the coupled mass, whereas recovery restores a load corresponding to approximately 12.9% of the mothership-only mass. These transitions are visible in the experimental mothership thrust record at the deployment and recovery events $\mathbf{C}_1$ and **F** in Fig. 9(a). The carrier must absorb these loading changes while maintaining stable altitude and planar station keeping, including immediately after capture when the two vehicles enter coupled descent.

Third, the two rotorcraft operate in close aerodynamic proximity during recovery. A momentum-theory estimate based on the installed propulsion configuration in Table 3 [41] gives characteristic hover-induced velocities of approximately 9.9 $\mathrm{m\,s^{-1}}$ for the child and 6.2 $\mathrm{m\,s^{-1}}$ for the mothership, with corresponding disk-loading scales of approximately 242 $\mathrm{N\,m^{-2}}$ and 93 $\mathrm{N\,m^{-2}}$, respectively. At the seated reference-point separation $g_{\min} = 0.40$ m, the vehicles therefore operate in a proximity regime in which parent–child rotor interference, lift variation, and asymmetric aerodynamic loading are physically relevant, as reported in prior experimental and computational studies [30, 28]. The present control architecture does not require an identified proximity-aerodynamics model; instead, these effects are treated as disturbances to be rejected by the closed-loop system. This constitutes an important distinction from landing on a rigid ground platform, for which the landing surface does not generate an interacting rotor flow field.

Fourth, the motion of the recovery platform is significant relative to the available capture region. The experimental capture radius is $r_{\mathrm{dock}} = 0.40$ m, whereas the mothership exhibited a planar station-keeping RMS error of 0.25 m under the reported outdoor recovery conditions. Thus, the characteristic carrier-motion scale is approximately 63% of the capture radius. Across the recovery campaign, the largest planar deviation observed within the capture envelope was 0.32 m. Terminal docking must therefore regulate the child relative to a target whose motion remains non-negligible relative to the allowable capture geometry.

Fifth, recovery depends on a distributed relative-state architecture. Both vehicles maintain their own RTK-enabled state estimates, and the child forms the terminal relative state using the mothership state received through the inter-vehicle link, together with its own onboard estimate. During the reported recovery campaign, the RTK solutions remained fixed, with receiver-reported horizontal uncertainty on the order of 0.03–0.06 m, whereas the measured mean planar alignment error at docking acceptance was approximately 0.180 m. These quantities are reported separately because receiver uncertainty and terminal radial alignment error are different statistical measures; no fraction of the observed alignment-error variance is inferred by squaring their ratio or through root-sum-of-squares subtraction. Instead, the experimental analysis evaluates the measured terminal relative-position statistics together with the recorded navigation and communication behavior.

Taken together, these characteristics explain why the recovery phase is the principal autonomy challenge of the system. The child must converge to an actively regulated airborne target, the carrier must accommodate deployment and recapture loading changes, both vehicles operate in aerodynamic proximity, terminal motion must remain within a finite thrust-and-tilt envelope, and the recovery logic must act on exchanged relative-state information while maintaining the prescribed close-proximity geometry. The objective of the proposed architecture is therefore not to introduce a new standalone controller, but to coordinate these requirements within a complete field-deployable recovery system.

This paper presents a complete rover–mothership–child autonomy framework designed for robust deployment, sortie execution, and autonomous recovery in outdoor environments. In the proposed architecture, the rover provides mission-level supervision and a stable communication anchor, the mothership UAV serves as both a transport vehicle and a mobile recovery platform, and the child UAV performs localized sensing or task execution before returning for recovery. The complete hardware system is implemented and experimentally evaluated in Section 8. Within this architecture, the recovery pipeline composes jerk-limited reference generation for smooth phase transitions [42, 43], disturbance-observer-augmented planar tracking to reject persistent external bias [31, 32], feasibility-aware vertical control consistent with thrust and tilt limits, and a discrete-time barrier-based safety filter [38, 39] that constrains the relative vertical geometry during close-proximity descent. To improve robustness under realistic communication conditions, the framework further incorporates redundant wireless links together with a lightweight prediction bridge [21] that maintains usable relative-state information during packet delays and dropouts.

The proposed system is implemented using ROS 2 for mission-level coordination and PX4 for onboard flight-critical stabilization, thereby preserving a clear separation between high-level autonomy and low-level vehicle control [44, 45]. The framework is evaluated in both control-oriented simulation and outdoor flight experiments using RTK-enabled GNSS-based state estimation. Across deployment, terminal docking, and synchronized descent, the results demonstrate stable tracking behavior, 17 successful recoveries in 20 outdoor attempts, quantitative terminal-alignment performance, and good consistency between simulated and experimental operation. Taken together, these results show that autonomous mothership recovery can be made sufficiently robust for realistic outdoor field use when disturbance rejection, feasibility constraints, safety filtering, and communication resilience are addressed within a unified system design.

### 1.2 | Contributions

The contributions of this work are stated at the system, integration, formalization, and experimental levels rather than as claims of novelty for the individual control algorithms.

First, at the *systems and experimental level*, this paper presents a complete field-validated rover–mothership–child autonomy architecture in which a ground rover provides supervision and communication support, a 14 kg hexacopter serves as the transport vehicle and hovering aerial recovery platform, and a 1.8 kg quadcopter performs an independent sortie before autonomously returning for aerial capture and synchronized descent. Relative to work that considers the same quadrotor-on-hexacopter configuration only in simulation [29], or investigates terminal landing on an airborne carrier as an isolated maneuver [28], the contribution here is the experimentally demonstrated integration of recovery into a complete heterogeneous mission sequence.

Second, at the *autonomy-integration level*, the paper specifies the interfaces through which jerk-bounded phase generation, disturbance-observer-augmented planar regulation, thrust-and-tilt feasibility projection, vertical safety filtering, and communication-aware carrier-state propagation operate as a unified recovery stack. The seventh-order phase generator follows established jerk-limited motion-generation methods [42, 43]; the planar disturbance observer follows established acceleration-mismatch and wind-rejection concepts [31, 32, 33]; and the carrier-state bridge follows prediction concepts used for delayed-state moving-platform landing [21]. The contribution is therefore their coordinated implementation and experimental evaluation under the specific requirements of a hovering aerial carrier rather than the individual algorithms themselves.

Third, at the *safety-formulation level*, the discrete-time barrier component is presented as a minimally invasive safety filter rather than as a new general control-barrier-function theory. The formulation is connected explicitly to established barrier-function methods [36, 38, 39, 40]. Section 5.3 states the nominal forward-invariance property, the admissible range and physical units of the barrier gain, the per-step actuation-feasibility condition, and the limitations associated with the relative-degree-two vertical geometry. This provides a formal justification for the safety mechanism while keeping the methodological claim appropriately scoped.

Fourth, at the *experimental level*, the recovery system is evaluated over 20 outdoor attempts, of which 17 resulted in successful docking. Rather than relying only on a representative trajectory, the experimental analysis reports aggregate and attempt-level terminal metrics, including alignment time, planar relative-position error at docking acceptance, maximum planar deviation within the terminal capture region, minimum relative vertical separation, navigation quality, and the behavior of the three unsuccessful attempts. The experimental results are interpreted as evidence for the performance of the integrated recovery architecture within the tested operating envelope rather than as evidence that any single sensing or control component alone determines capture accuracy.

Accordingly, no novelty is claimed for the individual building blocks of the recovery stack. The contribution of the paper is the formulation of hovering-carrier recovery as an integrated autonomy problem, the explicit composition and formal clarification of the required recovery mechanisms, and, most importantly, their validation on the complete rover–mothership–child robotic system in outdoor flight.

To provide context for the proposed autonomy framework, the next section describes the experimental mission sequence and operational workflow of the rover–mothership–child system, illustrating how the heterogeneous platforms coordinate during deployment, sortie execution, and recovery.

## 2 | System Architecture and Mission Profile

This paper considers a heterogeneous aerial–ground robotic team composed of three cooperative agents: a ground rover that provides mission supervision and a stable communication anchor, a hexacopter mothership UAV that transports the child vehicle and serves as a hovering recovery platform, and a quadcopter child UAV that executes localized sorties. Such heterogeneous UAV–UGV cooperation has been widely recognized as an effective strategy for combining the endurance, payload capacity, and logistical stability of ground vehicles with the agility and sensing reach of aerial vehicles [1, 2, 3, 5, 6]. The overall architecture and a representative outdoor mission sequence are shown in Fig. 2 and Fig. 3. Each aerial vehicle carries an RTK-capable GNSS receiver, an inertial measurement unit, and a barometric altimeter, which are fused onboard by the PX4 estimator. The mothership additionally carries a downward-facing one-dimensional lidar that provides an auxiliary close-range cue and supervisory monitoring signal during final recovery; the lidar does not enter the docking-acceptance decision. Both aerial vehicles operated with RTK-fixed navigation during the reported recovery campaign.

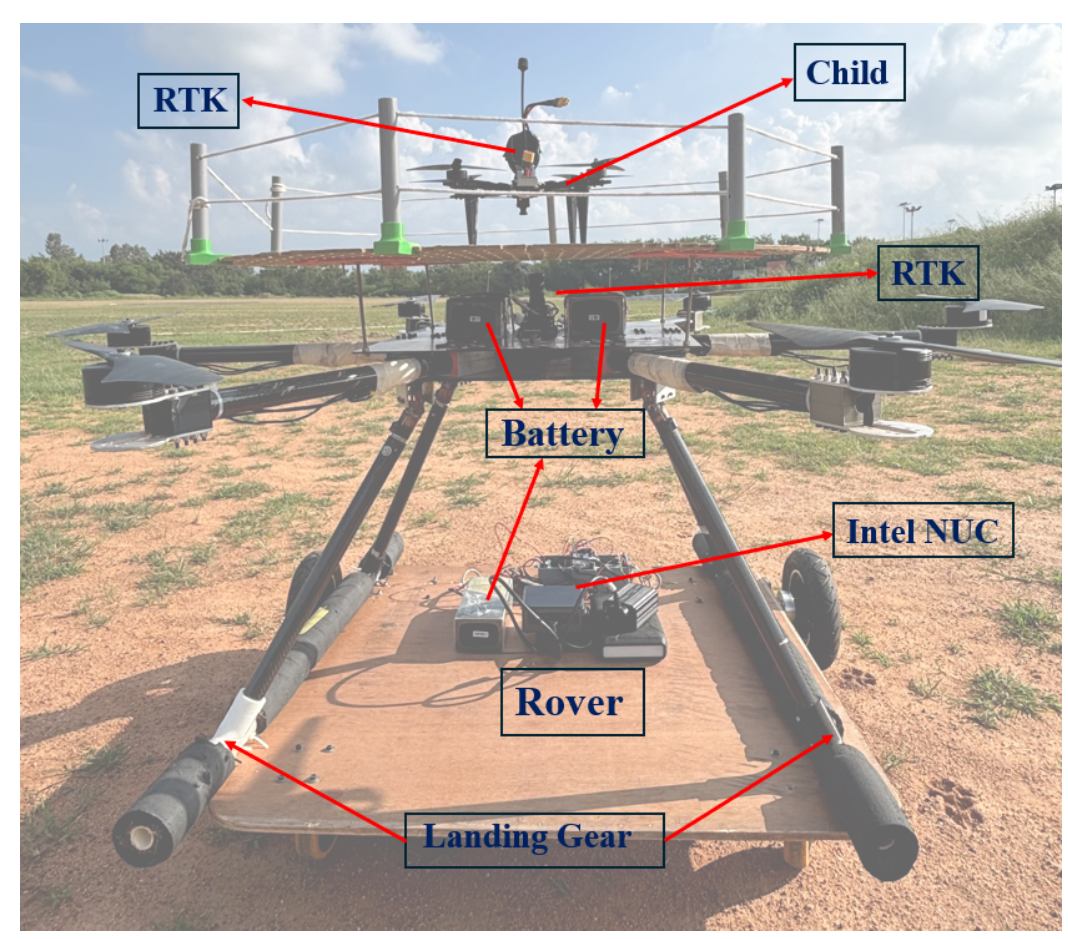

**FIGURE 1** | Hardware realization of the rover–mothership–child system used in outdoor experiments.

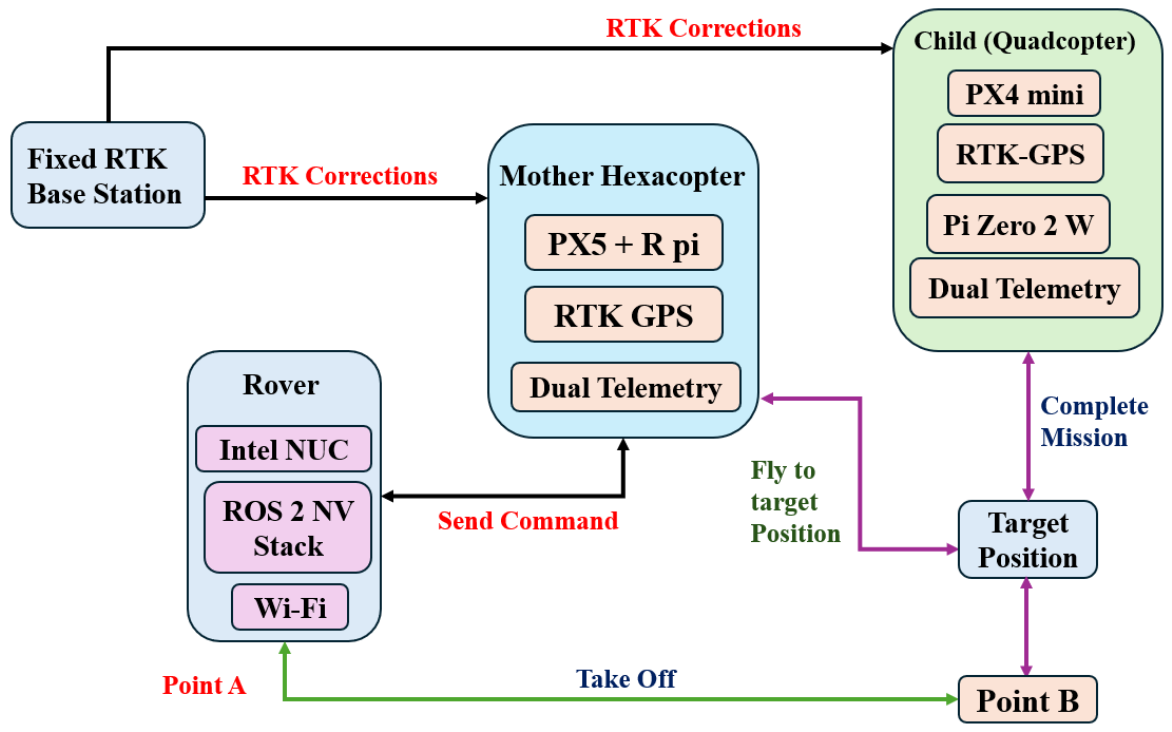

**FIGURE 2** | System architecture and mission workflow of the rover–mothership–child system.

At the system level, the key design principle is a strict separation between mission supervision and flight-critical stabilization. The rover executes sequencing, mission logic, navigation supervision, and operator interaction within a ROS 2-based supervisory layer, whereas both aerial vehicles perform onboard state estimation, attitude stabilization, and thrust control locally through PX4-based flight control [44, 45]. This division prevents supervisory computation and wireless-network variability from directly affecting inner-loop flight stability, while still enabling coordinated deploy–recover behavior through explicit command and state interfaces exchanged over the communication network.

The computational allocation follows the same partition. Mission sequencing and supervision are executed on the rover's onboard computer, whereas the recovery-oriented guidance and control pipeline of Sections 3–5 executes on the aerial-vehicle companion computers. This pipeline includes phase-reference generation, planar tracking with disturbance estimation, vertical regulation, feasibility projection, and the barrier-based safety filter. These recovery-level functions operate at the sampling period $T_s = 0.05$ s and provide setpoints to PX4, which closes the flight-critical attitude and thrust loops at higher internal rates. The platform assignment and autopilot interfaces are summarized in Table 3. This computational distinction is relevant to the discrete-time safety formulation because the outer-loop period $T_s$ defines the sampling interval used by the barrier condition in Section 5.3, including the admissibility requirement $\gamma T_s < 1$.

The mission organization exploits the operational strengths of the three platforms. The rover provides long-endurance ground mobility, persistent computation, a stable communications backbone, and a supervisory interface for mission authorization and monitoring. The mothership extends the effective reach of the child UAV by transporting it close to the region of interest, thereby reducing transit burden and preserving child-vehicle endurance for task-relevant operations. The child UAV is then deployed only for the local sortie segment, after which it returns to the mothership for autonomous recovery. This allocation of roles is consistent with broader marsupial, deployment–recovery, and carrier-supported mission concepts reported in both robotics and mothership-routing studies [6, 7, 11, 14].

Throughout this section, the system description follows the PX4 local navigation convention, with the world frame $\mathcal{W}$ defined according to North–East–Down (NED). This choice is consistent with onboard estimation, inter-vehicle communication, and logged flight data. The control-oriented development of Sections 3–5 is expressed in an equivalent $z$-up analysis frame, with the mapping between the implementation and analysis frames given explicitly in (6). Figures generated directly from the flight logs retain the NED convention and identify the corresponding frame in the axis labels or captions.

The overall mission workflow proceeds as follows. The rover first transports the system to the designated operating site, where the mothership initializes, arms, and performs autonomous takeoff. After reaching the designated operating region, the mothership deploys the child UAV to execute a local sortie. Once the sortie is complete, the child transitions to return-and-recovery behavior, performs a terminal rendezvous with the mothership, and docks onto the recovery platform. Following successful capture, the coupled aerial system performs a synchronized descent and landing. Figure 3 shows this experimentally executed single deployment–sortie–recovery mission sequence; it is not intended to represent a consecutive redeployment mission.

From a systems perspective, this workflow differs from a conventional single-UAV mission in two important ways. First, deployment and recovery are treated as integral components of the mission rather than as isolated initialization and termination events. Second, the mothership functions simultaneously as a transport vehicle, a docking base, and a dynamically coupled recovery platform. These roles impose requirements on communication structure, recovery logic, and safety-aware motion coordination that do not arise in standard point-to-point flight missions. They also introduce the two loading transitions identified in Section 1.1: release of the child during deployment and restoration of the child payload during recapture, both of which must be accommodated by the carrier's vertical control loop. Similar concerns appear in airborne docking, in-flight launch, and carrier-supported UAV operations, where the transition between independent flight and coupled recovery becomes a central systems challenge [18, 19, 16, 17]. The architecture in Fig. 2 is therefore designed to support not only nominal sortie execution, but also the close-proximity coordination needed for reliable terminal capture and synchronized descent under outdoor operating conditions.

## 2.1 | Mission phases

Operationally, the mission is organized as a sequence of well-defined phases so that task allocation, supervisory logic, and vehicle transitions remain explicit throughout the deployment–recovery cycle. Table 1 summarizes the principal phases together with their supervisory transition conditions. Each phase is associated with a distinct mission objective while remaining connected to the common motion-generation and recovery framework used throughout the system. This organization allows takeoff, transit, deployment, sortie execution, terminal rendezvous, capture, and synchronized descent to be coordinated explicitly without coupling the supervisory state machine to the flight-critical inner loops. This phase-based structure is especially important in heterogeneous aerial–ground systems, where different agents contribute distinct mobility, sensing, and logistical capabilities and where deployment and recovery must be treated as integral parts of the mission rather than as incidental initialization and termination events [6, 7].

**TABLE 1** | Mission phases and supervisory transition conditions. The transition from independent child flight to terminal recovery initiates close-proximity cooperative motion between the two aerial vehicles.

| # | Phase | Transition condition to next phase |
|---|---|---|
| 1 | Rover transit | System reaches the operating site; aerial vehicles remain inactive |
| 2 | Takeoff authorization | Rover grants authorization and the prescribed safety hold is completed |
| 3 | Climb and transit | Mothership reaches the designated operating region and loiter condition |
| 4 | Child sortie | Assigned local sortie is completed and return is commanded |
| 5 | Return and terminal approach | Terminal docking criterion is satisfied for $t_{\mathrm{dwell}}$ (capture), or the supervisory recovery criterion requests abandonment and return to loiter |
| 6 | Coupled hold and synchronized descent | Coupled aerial system reaches touchdown |

From the standpoint of recovery, the most critical transition occurs when the child exits the sortie phase and begins terminal return to the mothership. At this point, the mission logic shifts from independent task execution to close-proximity cooperative motion, in which relative geometry, communication continuity, and safe descent coordination become dominant requirements. The terminal logic therefore admits two supervisory outcomes: successful capture after satisfaction of the docking criterion over the prescribed dwell interval, or safe abandonment of an approach when the recovery conditions are not satisfied. Algorithm 1 defines these terminal decisions, while their experimental behavior is examined using the twenty-attempt recovery campaign in Section 8.

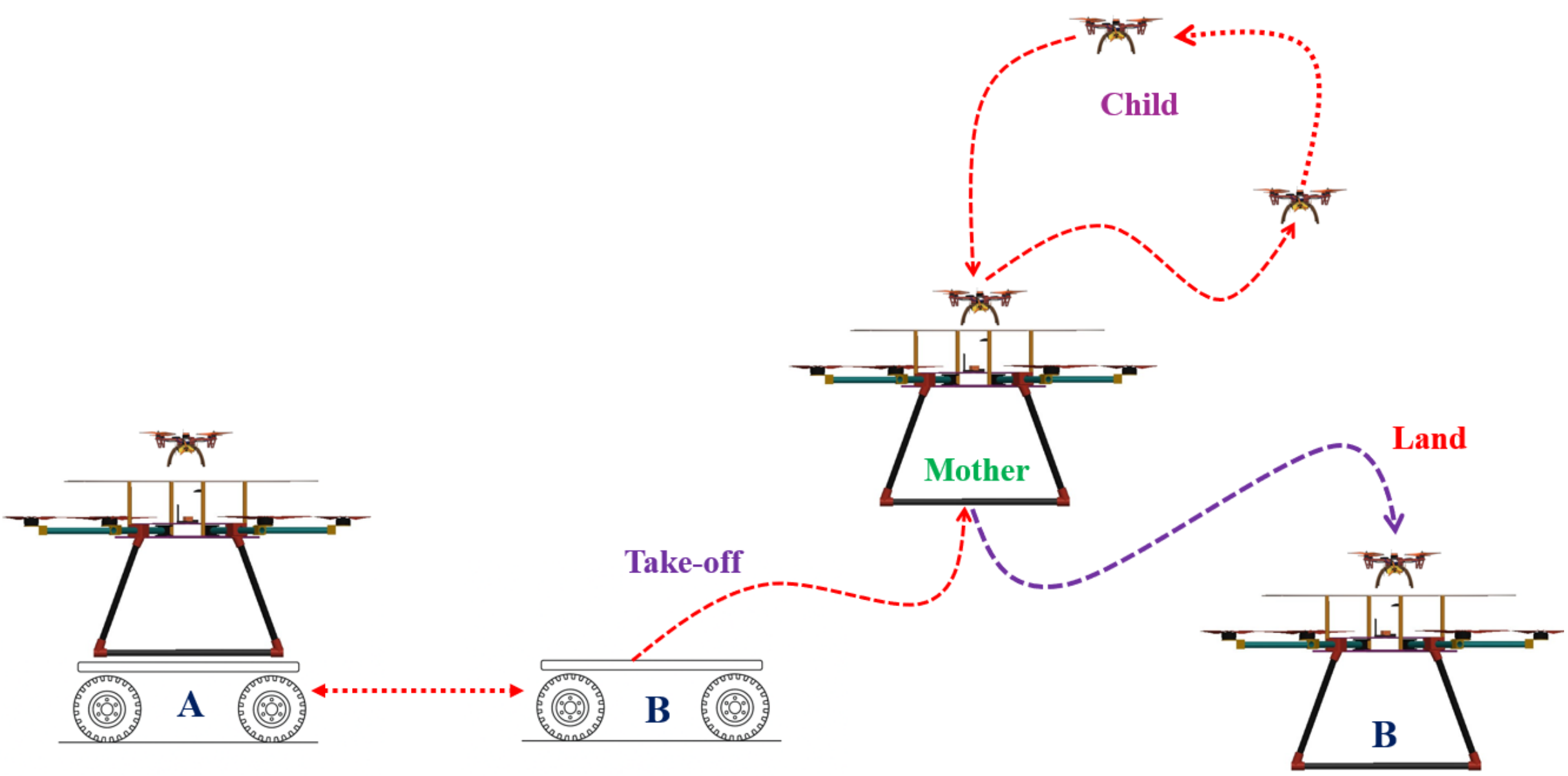

**FIGURE 3** | Representative experimental sequence of the rover–mothership–child system during outdoor operation.

## 2.2 | Jerk-bounded phase reference generation

To reduce excitation of unmodeled dynamics and improve repeatability across takeoff, terminal approach, docking, and synchronized descent, each mission phase is executed through a smooth point-to-point reference parameterization rather than by abrupt setpoint switching. Let the normalized phase time be defined as

$$u(t) = \mathrm{clip}\left(\frac{t}{T}, 0, 1\right), \tag{1}$$

where $T > 0$ denotes the nominal phase duration and $\mathrm{clip}(\cdot, 0, 1)$ saturates its argument to the interval $[0, 1]$. The phase is executed using the standard seventh-order time-scaling polynomial

$$s(u) = 35u^4 - 84u^5 + 70u^6 - 20u^7. \tag{2}$$

This polynomial satisfies $s(0) = 0$, $s(1) = 1$, and $s^{(j)}(0) = s^{(j)}(1) = 0$ for $j = 1, 2, 3$, yielding zero velocity, acceleration, and jerk at the endpoints. The reference is therefore continuous through jerk across phase boundaries when consecutive segments are joined at their endpoints. The normalized profile reaches 10% and 90% of its displacement at $u \approx 0.279$ and $u \approx 0.721$, respectively, giving a reference 10–90% rise time of approximately $0.443T$. This quantity is used later only as a reference-profile property when interpreting the simulated and measured transients.

Given an initial waypoint $\mathbf{p}_0$ and a terminal waypoint $\mathbf{p}_f$, the desired position and velocity references are generated as

$$\mathbf{p}_{\mathrm{des}}(t) = \mathbf{p}_0 + s\big(u(t)\big)\,(\mathbf{p}_f - \mathbf{p}_0), \tag{3}$$

$$\mathbf{v}_{\mathrm{des}}(t) = \dot{s}\big(u(t)\big)\,(\mathbf{p}_f - \mathbf{p}_0), \tag{4}$$

where

$$\dot{s}(u) = \frac{1}{T}\left(140u^3 - 420u^4 + 420u^5 - 140u^6\right). \tag{5}$$

This construction yields a jerk-bounded translational reference with smooth endpoint transitions, which is desirable for repeated autonomous operation in the presence of estimator noise, actuator bandwidth limits, and phase-dependent mission logic. In particular, using the same motion-generation structure across deployment, terminal approach, docking, and coupled descent provides a common dynamical grammar for the mission and reduces sensitivity to abrupt transients during phase changes. This viewpoint is consistent with jerk-aware trajectory-generation methods developed for real-time robotic motion planning and online reference generation [42, 43]. The resulting references $\{\mathbf{p}_{\mathrm{des}}, \mathbf{v}_{\mathrm{des}}\}$ define the nominal translational objectives subsequently tracked by the recovery-oriented controller except during terminal approach, where the horizontal velocity reference is replaced by the bounded relative-velocity funnel law of Section 5.4.

## 2.3 | Communication roles and redundancy

Reliable coordination during deployment and recovery requires robust exchange of supervisory commands, vehicle states, and timing information under realistic outdoor communication impairments. In practice, packet loss, variable latency, and intermittent link degradation are difficult to avoid during multi-vehicle operation, particularly when the aerial agents are mobile and the communication geometry changes throughout the mission. To improve resilience, the proposed system uses redundant wireless links with distinct functional roles.

Wi-Fi is employed primarily for mission-level supervision, including rover-to-mothership authorization, operator interaction, and other non-time-critical exchanges when coverage is available. In parallel, the telemetry link provides a more persistent channel for command-and-control, health monitoring, and time-sensitive state exchange between the mothership and child UAV during deployment, sortie execution, terminal rendezvous, and recovery. The exchanged mothership state comprises time-stamped position and velocity information used by the child recovery logic. Acceleration and future reference commands are not required as part of this inter-vehicle state message; the resulting assumption in the one-step vertical prediction is addressed explicitly in Section 5.3. This separation prevents non-critical supervisory traffic from interfering with the state exchange most relevant to safe close-proximity coordination.

The practical importance of this design is twofold. First, it reduces dependence on any single wireless channel and therefore improves robustness when one link experiences temporary degradation. Second, because low-level stabilization and state estimation remain onboard each aerial vehicle through the PX4-based flight-control stack, network variability does not directly destabilize the flight-critical inner loops [44, 45]. Instead, the communication network supports coordination at the supervisory and recovery levels, where redundancy, bounded-rate update logic, and prediction can be used to tolerate delay and packet loss without sacrificing basic vehicle stability.

This communication-role separation is particularly important during terminal recovery, where delayed or intermittent mothership-state information can degrade relative-motion regulation even when the individual aerial vehicles remain internally stable. By isolating flight-critical control onboard and reserving the inter-vehicle network for supervisory coordination and state exchange, the architecture provides a robust foundation for outdoor docking and synchronized descent.

Communication performance is evaluated explicitly in the later communication and experimental sections. The 3% packet-loss condition used for robustness testing is reported as a controlled communication-impairment condition handled by the prediction bridge and is distinguished from the nominal RTK-fixed field operation. The corresponding communication timing and recovery behavior are analyzed together with the experimental results rather than being treated as an unobserved limitation.

## 3 | Vehicle Dynamics and Actuation Model

The system description in Section 2 follows the PX4 local navigation convention, which is expressed in a North–East–Down frame for consistency in implementation and logging. For control analysis, however, it is convenient to use an equivalent inertial analysis frame $\mathcal{W}_u$ with $z$ positive upward. Let $\mathbf{p} = [x, y, z]^\top$ and $\mathbf{v} = \dot{\mathbf{p}}$ denote the vehicle position and velocity in $\mathcal{W}_u$. The logged PX4 states are mapped to this analysis frame by the vertical sign conversion

$$z_u = -z_{\mathrm{NED}}, \qquad v_{z,u} = -v_{z,\mathrm{NED}}, \qquad a_{z,u} = -a_{z,\mathrm{NED}}, \tag{6}$$

with the horizontal components unchanged. All control-oriented development in Sections 3–5 is expressed in $\mathcal{W}_u$. In particular, the relative vertical geometry used by the safety filter is evaluated after applying (6), so that a child vehicle physically above the mothership satisfies $z_{\mathrm{C}} > z_{\mathrm{M}}$. Figures reproduced directly from PX4 flight logs retain the NED convention and are identified accordingly. This separation preserves implementation fidelity while allowing the guidance and safety development to be expressed in a conventional $z$-up form.

Each UAV $u \in \{\mathrm{M}, \mathrm{C}\}$ is modeled at the outer-loop level by translational dynamics with gravity, realized thrust, and aerodynamic disturbance. The objective of this model is not to reproduce the full rigid-body dynamics of the aircraft, but rather to capture the dominant translational behavior relevant to phase transitions, docking, and synchronized descent. Inner-loop realization of commanded thrust direction and magnitude is represented by first-order dynamics, which is consistent with the practical PX4 architecture in which high-rate onboard attitude and thrust loops realize lower-rate outer-loop commands [45]. For the parameters used in this paper, the dominant attitude time constant $\tau_{\mathrm{att}} = 0.15$ s corresponds to a characteristic bandwidth of approximately 6.7 rad s$^{-1}$, which is well above the outer-loop frequencies implied by the gains of Table A3. The resulting separation supports the use of first-order actuation lags in the control-oriented model.

### 3.1 | Translational dynamics

The translational motion of each aerial vehicle is modeled as

$$m\ddot{\mathbf{p}} = \mathbf{F}_{\mathrm{thrust}} + \mathbf{F}_{\mathrm{drag}} + m\mathbf{g}, \qquad \mathbf{g} = [0, 0, -g]^\top, \quad g = 9.81\ \mathrm{m/s^2}, \tag{7}$$

where $m$ is the vehicle mass, $\mathbf{F}_{\mathrm{thrust}}$ is the realized thrust force, and $\mathbf{F}_{\mathrm{drag}}$ represents the net aerodynamic disturbance. Let $\mathbf{b}_{z,w} \in \mathbb{R}^3$ denote the realized thrust direction expressed in $\mathcal{W}_u$, with $\|\mathbf{b}_{z,w}\| = 1$, and let $T_{\mathrm{act}}$ denote the realized thrust magnitude. The corresponding thrust acceleration is

$$\mathbf{a}_{\mathrm{thrust}} = \frac{T_{\mathrm{act}}}{m}\,\mathbf{b}_{z,w}. \tag{8}$$

This form is sufficiently expressive for outer-loop guidance and safety design, while keeping the model compact enough for simulation, controller synthesis, and implementation-level interpretation. Aerodynamic effects represented by $\mathbf{F}_{\mathrm{drag}}$ are treated at this level as external translational disturbances; their control-oriented simulation model is introduced separately in Section 4.2.

### 3.2 | Inner-loop realization as first-order lags

The low-level realization of thrust direction and thrust magnitude is approximated by first-order lag dynamics,

$$\dot{\mathbf{b}}_{z,w} = \frac{1}{\tau_{\mathrm{att}}}(\mathbf{b}_{z,\mathrm{des}} - \mathbf{b}_{z,w}), \tag{9}$$

$$\dot{T}_{\mathrm{act}} = \frac{1}{\tau_T}(T_{\mathrm{cmd}} - T_{\mathrm{act}}), \tag{10}$$

where $\mathbf{b}_{z,\mathrm{des}}$ is the commanded thrust direction, $T_{\mathrm{cmd}}$ is the commanded thrust magnitude, and $\tau_{\mathrm{att}}$ and $\tau_T$ denote the dominant attitude and thrust response time constants. In simulation, $\mathbf{b}_{z,w}$ is renormalized after numerical integration so that the unit norm is preserved. This reduced-order representation captures the dominant delay between outer-loop commands and realized motion without introducing unnecessary inner-loop complexity, and it is adequate for reproducing the phase-transition behavior observed in both simulation and outdoor operation.

The thrust lag in (10) is also relevant to the one-step prediction used by the discrete-time safety filter. For a commanded acceleration change of magnitude $|\Delta a|$, the position discrepancy over one control interval that results from replacing the first-order thrust response by instantaneous acceleration is

$$\left|\Delta a\right|\left[\tau_T T_s + \tau_T^2\left(e^{-T_s/\tau_T} - 1\right)\right]. \tag{11}$$

With $T_s = 0.05$ s and $\tau_T = 0.10$ s, this becomes

$$\left|\Delta a\right|\left[\tau_T T_s + \tau_T^2\left(e^{-T_s/\tau_T} - 1\right)\right] = 1.07 \times 10^{-3}\left|\Delta a\right|, \tag{12}$$

which is used later only to characterize the modeling error of the one-step vertical prediction in Section 5.3.

### 3.3 | Feasibility constraints and command projection

During terminal recovery and synchronized descent, the outer loop must avoid requesting accelerations that cannot be realized by the vehicle because of thrust and tilt limitations. Let the planar command be $\mathbf{a}_{\mathrm{cmd},xy} = [a_{\mathrm{cmd},x}, a_{\mathrm{cmd},y}]^\top$ and the vertical command be $a_{\mathrm{cmd},z}$. The corresponding commanded specific force is written as

$$\mathbf{f} = \begin{bmatrix} a_{\mathrm{cmd},x} \\ a_{\mathrm{cmd},y} \\ g + a_{\mathrm{cmd},z} \end{bmatrix} = \begin{bmatrix} \mathbf{f}_{xy} \\ f_z \end{bmatrix}. \tag{13}$$

Feasibility requires that

$$\|\mathbf{f}\| \le f_{\max} = \frac{T_{\max}}{m}, \tag{14}$$

where $T_{\max}$ is the maximum thrust made available to the outer-loop command layer and $\theta_{\max}$ is the admissible tilt angle. Accordingly, $f_{\max}$ is a software-enforced control-shell limit rather than a claim about the absolute propulsion-system maximum. The physical platform parameters are summarized separately in Table 3.

The tilt definition must remain valid when the commanded vertical specific force approaches zero. The single-argument inverse tangent of $\|\mathbf{f}_{xy}\|/f_z$ is undefined at $f_z = 0$ and gives an incorrect geometric interpretation for $f_z < 0$. The commanded tilt is therefore defined using the four-quadrant inverse tangent,

$$\theta = \mathrm{atan2}\left(\|\mathbf{f}_{xy}\|, f_z\right) \in [0, \pi], \qquad \theta \le \theta_{\max}. \tag{15}$$

For the non-inverted operating envelope considered here, $\theta_{\max} < \pi/2$, so (15) requires $f_z > 0$ and

$$\|\mathbf{f}_{xy}\| \le f_z \tan\theta_{\max}. \tag{16}$$

A strictly positive lower bound is therefore imposed on the vertical specific force before the tilt projection is evaluated,

$$f_z \leftarrow \max\left(f_z, f_z^{\min}\right), \qquad f_z^{\min} > 0, \tag{17}$$

which is equivalent to the commanded-descent constraint

$$a_{\mathrm{cmd},z} \ge f_z^{\min} - g. \tag{18}$$

This prevents the outer loop from requesting a zero-thrust or thrust-reversed vertical condition during synchronized descent.

To enforce these limits, the command is projected to the admissible set in two steps. First, the planar component is scaled, if necessary, so that the tilt bound is satisfied:

$$\mathbf{f}_{xy} \leftarrow \begin{cases} \min\left\{1, \dfrac{f_z \tan\theta_{\max}}{\|\mathbf{f}_{xy}\|}\right\}\mathbf{f}_{xy}, & \|\mathbf{f}_{xy}\| > 0, \\ \mathbf{0}, & \|\mathbf{f}_{xy}\| = 0. \end{cases} \tag{19}$$

The explicit zero-command case avoids the formal 0/0 ambiguity present when the projection is written only as a scaling ratio.

Second, the planar component is reduced, if necessary, so that the thrust bound is satisfied with the vertical specific force held fixed:

$$\mathbf{f}_{xy} \leftarrow \begin{cases} \min\left\{1, \dfrac{\sqrt{\max\left(0, f_{\max}^2 - f_z^2\right)}}{\|\mathbf{f}_{xy}\|}\right\}\mathbf{f}_{xy}, & \|\mathbf{f}_{xy}\| > 0, \\ \mathbf{0}, & \|\mathbf{f}_{xy}\| = 0. \end{cases} \tag{20}$$

The order of these operations is important. Scaling the complete specific-force vector to enforce the thrust limit would also reduce $f_z$ and could violate the lower bound imposed by (17). Equation (20) instead preserves the vertical specific force and assigns any required reduction to the planar command. Consequently, the vertical-force floor is preserved, and reducing $\|\mathbf{f}_{xy}\|$ at fixed $f_z$ cannot increase the commanded tilt. The construction is feasible whenever $f_z \le f_{\max}$. If $f_z > f_{\max}$, the requested vertical specific force alone exceeds the outer-loop actuation envelope and no admissible command exists; this condition is passed to the supervisory layer and is treated consistently with the safety-filter feasibility condition in Section 5.3.

For later use by the vertical safety filter, the admissible vertical-acceleration interval can be written explicitly once $\mathbf{f}_{xy}$ is fixed. Combining (14), (16), and (17) gives

$$\begin{aligned} a_z &\in \left[a_z^{\min}, a_z^{\max}\right], \\ a_z^{\max} &= \sqrt{f_{\max}^2 - \|\mathbf{f}_{xy}\|^2} - g, \\ a_z^{\min} &= \max\left\{\frac{\|\mathbf{f}_{xy}\|}{\tan\theta_{\max}}, f_z^{\min}\right\} - g. \end{aligned} \tag{21}$$

The interval is non-empty if and only if $a_z^{\min} \le a_z^{\max}$, that is,

$$\max\left\{\frac{\|\mathbf{f}_{xy}\|}{\tan\theta_{\max}}, f_z^{\min}\right\} \le \sqrt{f_{\max}^2 - \|\mathbf{f}_{xy}\|^2}. \tag{22}$$

The first branch of (22) reduces to

$$\|\mathbf{f}_{xy}\| \le f_{\max}\sin\theta_{\max}, \tag{23}$$

which bounds the planar demand that can be satisfied simultaneously with the thrust and tilt constraints. The second branch requires the available thrust to remain sufficient to maintain the imposed vertical-force floor. If (22) is violated, priority is given to vertical authority, and the planar command is reduced until a non-empty vertical-acceleration interval is recovered. This same vertical-priority convention is used by the barrier-based safety filter in Section 5.3.

For the implemented parameters, these constraints function primarily as safety backstops during nominal recovery. The mothership propulsion system provides a static thrust-to-weight ratio of approximately 3.6 at the 14.0 kg all-up mass (Table 3), while the conservative outer-loop limits are used by both vehicles. With $\theta_{\max} = 25°$, the tilt constraint would become active only for commanded vertical accelerations below approximately $-5.5\ \mathrm{m\,s^{-2}}$ for the mothership and $-3.4\ \mathrm{m\,s^{-2}}$ for the child. These values lie well below the vertical accelerations generated by the nominal recovery trajectories. Equivalently, the maximum planar commands correspond to hover-equilibrium tilt angles of approximately 11.5° and 17.0°, respectively. The logged nominal recovery commands remained inside this feasibility envelope, so the projection acts primarily as a safeguard against off-nominal combinations of planar and vertical demand rather than as the mechanism responsible for the nominal tracking performance.

This feasibility projection is especially important during safety-critical phases because it ensures that robustness and safety corrections remain compatible with the physical actuation limits of the platform.

# 4 | Simulation Disturbance and Measurement Model

The outdoor recovery problem is shaped by environmental disturbances and imperfect state information. In the simulation study, these effects are represented through wind, aerodynamic drag, and intermittent relative-state updates. These models are introduced to stress the autonomy stack under operating conditions representative of the outdoor experiments and to evaluate the response of the recovery architecture during deployment, terminal approach, docking, and synchronized descent.

The models in this section are therefore control-oriented simulation abstractions rather than replacements for the sensing and communication architecture used in flight. In the outdoor experiments, vehicle state is obtained from the RTK-enabled PX4 estimation pipelines described in Section 8, while the ambient wind condition for the test period is taken from the corresponding site-level government environmental record rather than from an onboard airspeed sensor.

## 4.1 | Wind disturbance

To represent the dominant atmospheric effects encountered during outdoor operation, the wind field is modeled as the sum of a steady background component and a stochastic gust component:

$$\mathbf{v}_{\text{wind}} = \mathbf{v}_{\text{steady}} + \mathbf{v}_{\text{gust}}, \tag{24}$$

where $\mathbf{v}_{\text{steady}}$ represents the slowly varying background wind and $\mathbf{v}_{\text{gust}}$ represents the fluctuating component. In the representative simulation, the magnitude of the steady component is selected to be consistent with the approximately $1.8\ \mathrm{m\,s^{-1}}$ ambient condition reported for the outdoor test period, while its direction is specified by the simulation scenario in Table A1. The stochastic term is then superimposed to evaluate sensitivity to short-duration fluctuations around this background condition.

The gust component is represented as a zero-mean, first-order stationary Gauss–Markov (Ornstein–Uhlenbeck) process,

$$d\mathbf{v}_{\text{gust}} = -\frac{1}{\tau_{\text{gust}}}\mathbf{v}_{\text{gust}}\,dt + \sigma_w\,d\mathbf{W}(t), \tag{25}$$

where $\mathbf{W}(t)$ is a vector standard Wiener process, $\tau_{\text{gust}} > 0$ is the gust correlation time, and $\sigma_w$ is the continuous-time driving-noise intensity. Writing the model as a stochastic differential equation removes the dimensional ambiguity associated with treating Gaussian white noise as an ordinary forcing term. The stationary process has

$$\mathbb{E}[\mathbf{v}_{\text{gust}}] = \mathbf{0}, \qquad \Sigma_{\text{ss}} = \frac{\sigma_w^2 \tau_{\text{gust}}}{2} I, \tag{26}$$

and autocovariance

$$R_{\text{gust}}(\Delta t) = \Sigma_{\text{ss}} \exp\left(-\frac{|\Delta t|}{\tau_{\text{gust}}}\right). \tag{27}$$

For implementation, the model is parameterized by the desired stationary standard deviation $\sigma_{\text{gust}}$, giving the following.

$$\sigma_w = \sigma_{\text{gust}}\sqrt{\frac{2}{\tau_{\text{gust}}}}, \tag{28}$$

and the initial gust state is sampled as

$$\mathbf{v}_{\text{gust}}[0] \sim \mathcal{N}\left(\mathbf{0}, \sigma_{\text{gust}}^2 I\right), \tag{29}$$

so that the stochastic process begins in its stationary distribution and does not introduce an artificial initialization transient. At the simulation sampling period $T_s$, the exact discrete-time realization is

$$\begin{aligned} \mathbf{v}_{\text{gust}}[k+1] &= a_g \mathbf{v}_{\text{gust}}[k] + \sqrt{1 - a_g^2}\,\sigma_{\text{gust}}\,\boldsymbol{\xi}[k], \\ a_g &= e^{-T_s/\tau_{\text{gust}}}, \qquad \boldsymbol{\xi}[k] \sim \mathcal{N}(0, I), \end{aligned} \tag{30}$$

which preserves the stationary variance and correlation time of the continuous process exactly at the sampling instants.

Although simplified, the model provides a useful control-oriented representation of the mild outdoor wind conditions considered in the present study [31, 33].

The selected gust model primarily stresses rejection of low-frequency lateral disturbance rather than severe turbulence. The vertical wind component is set to zero in the simulation scenario so that the principal comparison isolates horizontal docking regulation under the mild-wind operating condition represented in the field campaign. Consequently, the wind simulation should be interpreted as a controlled disturbance-rejection test around the experimentally relevant operating condition, rather than as a complete atmospheric-turbulence model.

## 4.2 | Aerodynamic drag

Aerodynamic drag is modeled as a quadratic function of the vehicle air-relative velocity [31, 32]. Defining

$$\mathbf{v}_{\text{rel}} = \mathbf{v} - \mathbf{v}_{\text{wind}}, \tag{31}$$

The corresponding control-oriented drag acceleration is separated into planar and vertical components as

$$\mathbf{a}_{\text{drag},xy} = -k_{\text{drag},xy} \|\mathbf{v}_{\text{rel},xy}\| \, \mathbf{v}_{\text{rel},xy}, \tag{32}$$

$$a_{\text{drag},z} = -k_{\text{drag},z} \, v_{\text{rel},z} |v_{\text{rel},z}|, \tag{33}$$

where $k_{\text{drag},xy}$ and $k_{\text{drag},z}$ are the effective drag coefficients in the horizontal and vertical channels, respectively, expressed per unit mass so that (32)–(33) return accelerations directly.

This form provides a compact approximation of the dominant velocity-dependent aerodynamic effects without introducing unnecessary model complexity. In particular, it captures the increase in disturbance magnitude with air-relative speed and therefore provides a useful mechanism for evaluating tracking degradation during outbound motion, return, and terminal recovery under wind.

The model is intentionally isotropic in the horizontal plane and does not attempt to resolve attitude-dependent aerodynamic coefficients or the detailed rotor–rotor interaction that occurs when the two vehicles enter close proximity. Prior work cited in Section 1.1 shows that such interaction can modify aerodynamic loading near an aerial carrier [30, 28]; in this paper, those proximity effects are treated as disturbances to the closed-loop system rather than as an identified interference model. This choice keeps the simulation focused on evaluating the implemented recovery architecture instead of introducing an aerodynamic model that was not used by the flight controller.

Equations (32)–(33) therefore define the simulation disturbance model. Where a motion-correlated compensation term is used in the implemented vertical controller, it is formed from the onboard inertial velocity estimate as described separately in Section 5.2; the controller does not require a direct measurement of air-relative velocity.

### 4.3 | Intermittent relative-state updates

Terminal recovery uses the relative state formed from the child's onboard RTK-enabled state estimate and the time-stamped mothership state received over the inter-vehicle communication link. In simulation, the same interface is represented by an intermittent noisy relative-state update so that the recovery logic can be exercised under temporary update loss without changing the downstream guidance law. The model therefore represents the availability and uncertainty of the relative information delivered to the recovery logic, rather than a separate physical relative-navigation sensor.

When a valid update is available, the simulated relative-position measurement is

$$\Delta\mathbf{p}_{\text{meas}}[k] = \left(\mathbf{p}_{\text{M}}[k] - \mathbf{p}_{\text{C}}[k]\right) + \mathbf{n}_{\text{rel}}[k], \tag{34}$$

with

$$\mathbf{n}_{\text{rel}}[k] \sim \mathcal{N}\left(\mathbf{0}, \operatorname{diag}\left(\sigma_{xy}^2, \sigma_{xy}^2, \sigma_z^2\right)\right), \tag{35}$$

and an availability variable

$$\delta_{\text{rel}}[k] \in \{0, 1\}, \tag{36}$$

where $\delta_{\text{rel}}[k] = 1$ indicates that a new relative-state update is available and $\delta_{\text{rel}}[k] = 0$ indicates that no new update is delivered at that sampling instant. Separating the validity variable from the measurement value avoids the misleading interpretation that a lost update produces a measured displacement of zero.

The estimate passed to the terminal guidance law is updated according to

$$\Delta\hat{\mathbf{p}}[k] = \begin{cases} \Delta\mathbf{p}_{\text{meas}}[k], & \delta_{\text{rel}}[k] = 1, \\ \Delta\hat{\mathbf{p}}[k-1], & \delta_{\text{rel}}[k] = 0, \end{cases} \tag{37}$$

before the communication-prediction mechanism of Section 6 is applied. In other words, a missed update preserves the last valid relative information rather than presenting the controller with a false zero-error condition. The elapsed time since the most recent valid update is tracked explicitly and is used by the recovery logic together with the prediction bridge to determine whether terminal guidance may continue or whether the approach should be safely abandoned.

The anisotropic covariance allows the horizontal and vertical measurement qualities to differ, which is often more realistic than assuming isotropic close-range sensing [23, 27].

The parameters $\sigma_{xy}$, $\sigma_z$, and the update-availability settings used in simulation are reported in Table A1. They are simulation-test parameters and are kept conceptually distinct from the receiver-reported RTK accuracy and communication statistics reported for the outdoor experiments. This separation prevents simulated sensor/update uncertainty from being interpreted as a measured field error source.

## 5 | Hierarchical Control and Safety

The proposed recovery stack combines nominal trajectory tracking with explicit robustness and safety mechanisms. Planar motion is regulated by an outer-loop proportional-derivative controller augmented with a lightweight disturbance observer for rejection of low-frequency aerodynamic bias. Vertical motion is regulated by a proportional-derivative-integral structure with feasibility-aware command generation. During close-proximity operation, a discrete-time barrier mechanism modifies the child vehicle's vertical command whenever the predicted relative geometry threatens the minimum allowable separation. This layered structure preserves nominal performance when the vehicles are well separated while introducing progressively stronger constraint-aware behavior as docking and synchronized descent begin.

The three mechanisms have distinct primary roles. The disturbance observer compensates slowly varying planar acceleration mismatch, the feasibility projection prevents the outer loop from requesting commands outside the available thrust-and-tilt envelope, and the barrier-based safety filter modifies the vertical command only when the predicted relative geometry approaches its prescribed boundary. The individual mechanisms are based on established control concepts; their role here is their coordinated implementation within the aerial-recovery architecture.

### 5.1 | Planar tracking with disturbance observer

Let

$$\mathbf{e}_{xy} = \mathbf{p}_{xy}^{\text{des}} - \mathbf{p}_{xy}, \qquad \mathbf{e}_{v,xy} = \mathbf{v}_{xy}^{\text{des}} - \mathbf{v}_{xy}, \tag{38}$$

denote the planar position and velocity tracking errors. Following the outer-loop PD acceleration-reference structure used by Smeur et al. [33], the nominal planar command is

$$\mathbf{a}_{xy}^{\mathrm{PD}} = K_{p,xy}\mathbf{e}_{xy} + K_{d,xy}\mathbf{e}_{v,xy}. \quad (39)$$

For the gains in Table A3, the planar-loop bandwidth remains below the inner-loop attitude bandwidth, while terminal closing speed is constrained separately by the docking guidance law in Section 5.4.

To compensate for slowly varying drag and crosswind effects, we use a lightweight discrete disturbance estimator. Smeur et al. [33] employ finite-difference reconstruction and filtering of acceleration-related signals. Using the same principle, the measured planar acceleration is reconstructed as

$$\hat{\mathbf{a}}_{xy}[k] = \frac{\mathbf{v}_{xy}[k] - \mathbf{v}_{xy}[k-1]}{T_s}, \quad (40)$$

and low-pass filtered as

$$\tilde{\mathbf{a}}_{xy}[k] = (1-\alpha_\ell)\tilde{\mathbf{a}}_{xy}[k-1] + \alpha_\ell \hat{\mathbf{a}}_{xy}[k]. \quad (41)$$

The filtering is also consistent with Waslander and Wang [31], who treat wind disturbances as low-bandwidth variations relative to the onboard measurement rate.

Waslander and Wang [31] further relate deviations between expected and measured acceleration to aerodynamic disturbances, while Ha and Park [32] explicitly estimate and compensate for translational disturbances. Motivated by these ideas, we use the following simplified discrete residual observer:

$$\hat{\mathbf{d}}_{xy}[k] = (1-\alpha_d)\hat{\mathbf{d}}_{xy}[k-1] + \alpha_d\Big(\tilde{\mathbf{a}}_{xy}[k] - \mathbf{a}_{\mathrm{cmd},xy}[k-1]\Big), \quad (42)$$

where $\alpha_\ell, \alpha_d \in (0,1)$. The one-sample delay preserves causality by comparing the realized acceleration with the command applied during the preceding sampling interval.

Following the disturbance-cancellation principle of Ha and Park [32], the estimated disturbance is subtracted from the nominal acceleration command:

$$\mathbf{a}_{\mathrm{cmd},xy} = \mathrm{sat}_{a_{xy}^{\max}}\left(\mathbf{a}_{xy}^{\mathrm{PD}} - \hat{\mathbf{d}}_{xy}\right), \quad (43)$$

where

$$\mathrm{sat}_{a_{xy}^{\max}}(\mathbf{r}) = \begin{cases} \min\left\{1, \dfrac{a_{xy}^{\max}}{\|\mathbf{r}\|}\right\}\mathbf{r}, & \|\mathbf{r}\| > 0, \\ \mathbf{0}, & \|\mathbf{r}\| = 0. \end{cases} \quad (44)$$

Equation (44) is an implementation-level definition used here to bound the planar acceleration magnitude.

The observer, therefore estimates only the slowly varying mismatch between commanded and realized planar acceleration, rather than the complete wind field. This is sufficient to suppress persistent lateral Bias during terminal docking while keeping the commanded acceleration within the prescribed limit.

## 5.2 | Vertical regulation with feasibility-aware command generation

Vertical motion is regulated using a proportional-integral-derivative command law,

$$e_z[k] = z_{\mathrm{des}}[k] - z[k], \qquad e_{v,z}[k] = v_z^{\mathrm{des}}[k] - v_z[k], \quad (45)$$

$$I_z[k+1] = \mathrm{clip}\big(I_z[k] + T_s e_z[k],\, I_z^{\min},\, I_z^{\max}\big), \quad (46)$$

$$a_{\mathrm{cmd},z} = K_{p,z}e_z + K_{d,z}e_{v,z} + K_{i,z}I_z - \hat{a}_{\mathrm{drag},z}, \quad (47)$$

where the drag-compensation term adopts the velocity-dependent functional form of (33), and the integral state is bounded using the limits to prevent excessive accumulation during constrained vertical motion.

The flight controller does not require an air-relative velocity measurement. Instead, the vertical compensation term is evaluated from the fused inertial vertical-velocity estimate supplied by the PX4 estimator,

$$\hat{a}_{\mathrm{drag},z} = -k_{\mathrm{drag},z}\, v_{\mathrm{EKF},z}\left|v_{\mathrm{EKF},z}\right|, \quad (48)$$

where $v_{\mathrm{EKF},z}$ is the vertical velocity in the analysis frame obtained from the onboard estimator that fuses RTK-enabled GNSS, inertial, and barometric information. Equation (48) is therefore a motion-correlated compensation term rather than a measurement or estimate of true aerodynamic drag. In simulation, by contrast, the aerodynamic disturbance itself is generated from the modeled air-relative velocity according to Section 4.2.

Any discrepancy between inertial-velocity-based compensation and the actual vertical aerodynamic loading is treated as residual model mismatch and is accommodated by the bounded integral action rather than by an explicit wind estimator.

The asymmetry between the planar and vertical channels is deliberate. In the horizontal plane, the disturbance observer estimates the realized acceleration mismatch directly and therefore absorbs crosswind, drag, and low-frequency modeling errors without assigning them to individual aerodynamic mechanisms. In the vertical channel, the feedforward term (48) removes the motion-correlated component while the bounded integral state compensates persistent residual error. Bounding $I_z$ is particularly important after capture, when the child remains armed while mechanically constrained by the recovery plate; without anti-windup, persistent position error against the contact constraint could accumulate and later generate an unnecessarily large vertical command.

The vertical command is not used in isolation; instead, it is combined with the planar command through the specific-force representation in (13) and subsequently projected to the admissible thrust-and-tilt set defined by (14)–(21). This feasibility-aware construction is particularly important near docking and during coupled descent, where large vertical corrections could otherwise drive the vehicle into excessive tilt or thrust saturation and thereby degrade controllability.

## 5.3 | Discrete-time barrier-based safety filter for vertical geometry

During terminal approach, docking, and synchronized descent, the child vehicle is required to respect the minimum relative vertical geometry of the recovery configuration. Let the barrier

function be

$$h[k] = \big(z_{\mathrm{C}}[k] - z_{\mathrm{M}}[k]\big) - g_{\min}, \tag{49}$$

so that

$$\mathcal{C} = \{\mathbf{x} : h(\mathbf{x}) \geq 0\}. \tag{50}$$

$g_{\min}$ is the vertical offset between the state-estimation reference points of the two vehicles when the child is fully seated on the recovery plate. It therefore represents the seated geometry rather than a required pre-contact clearance. The safety filter does not prevent capture; instead, it prevents the commanded relative vertical motion from driving the two vehicles below their prescribed seated geometry while both aerial vehicles remain actuated. Before seating, $h > 0$, and the filter limits how rapidly the child may close the remaining vertical separation. After seating, the recovery structure mechanically constrains the two vehicles near $h = 0$; the filter remains active as a command safeguard so that the child controller does not request further reduction of the relative vertical geometry during the coupled phase.
A discrete-time barrier condition is enforced in the form

$$h[k+1] \geq (1-\gamma T_s)h[k], \qquad \gamma \in \left(0, \frac{1}{T_s}\right), \tag{51}$$

The parameter $\gamma$ has units of $\mathrm{s}^{-1}$. Because $\gamma T_s \in (0,1)$, the factor $(1-\gamma T_s)$ lies strictly between zero and one and specifies the maximum permitted decay of the barrier value over one sampling interval. This is the discrete-time barrier form used here, following established discrete-time barrier formulations [38] and is consistent with the forward-invariance interpretation of continuous-time control barrier functions [36]. For the implemented values $T_s = 0.05$ s and $\gamma = 3.0\ \mathrm{s}^{-1}$, $\gamma T_s = 0.15$.

Proposition 1 (nominal discrete-time forward invariance). *Suppose $h[0] \geq 0$, the condition (51) is satisfied at every sampling instant, $\gamma \in (0, 1/T_s)$, and the one-step state prediction used to enforce the constraint is exact. Then*

$$h[k] \geq (1-\gamma T_s)^k h[0] \geq 0$$

*for all $k \geq 0$. Consequently, the safe set $\mathcal{C}$ is forward invariant.*
*Proof.* Since $\gamma T_s \in (0,1)$, $1-\gamma T_s > 0$. Thus, if $h[k] \geq 0$, then

$$h[k+1] \geq (1-\gamma T_s)h[k] \geq 0.$$

Repeated application gives $h[k] \geq (1-\gamma T_s)^k h[0] \geq 0$, which proves the result by induction.

Proposition 1 states the nominal property of the implemented filter. The two practical requirements behind this result—one-step prediction accuracy and per-step actuation feasibility—are treated explicitly below rather than being implicit assumptions.
The child vehicle's safe vertical command is obtained by solving the scalar optimization problem

$$\min_{a_{\mathrm{safe},z}} \left(a_{\mathrm{safe},z} - a_{\mathrm{cmd},z}\right)^2 \qquad \text{s.t.} \qquad \hat{h}[k+1] \geq (1-\gamma T_s)h[k], \tag{52}$$

together with the feasible vertical-acceleration interval (21) imposed by the thrust-and-tilt limits.
Under the one-step constant-acceleration prediction,

$$\hat{h}[k+1] = h[k] + T_s\big(v_{\mathrm{C},z}[k] - v_{\mathrm{M},z}[k]\big) + \frac{T_s^2}{2}\big(a_{\mathrm{safe},z}[k] - a_{\mathrm{M},z}^{\mathrm{ref}}[k]\big), \tag{53}$$

where $a_{\mathrm{M},z}^{\mathrm{ref}}$ denotes the mothership vertical reference acceleration.
The inter-vehicle state message contains the mothership position and velocity, but not a future acceleration reference. During terminal alignment, the mothership is commanded to hold station, so the implementation uses

$$a_{\mathrm{M},z}^{\mathrm{ref}} \equiv 0 \tag{54}$$

in the one-step safety prediction. During the subsequent smooth coupled descent, any mismatch introduced by this approximation is interpreted as prediction error rather than as a separately communicated control input.
Because (53) is affine in the scalar decision variable, the optimization has a closed-form solution. Define

$$\Delta v[k] = v_{\mathrm{C},z}[k] - v_{\mathrm{M},z}[k]. \tag{55}$$

With (54),

$$a_z^{\mathrm{req}}[k] = -\frac{2}{T_s}\big(\gamma h[k] + \Delta v[k]\big), \tag{56}$$

$$a_{\mathrm{safe},z}[k] = \mathrm{clip}\big(\max\{a_{\mathrm{cmd},z}[k], a_z^{\mathrm{req}}[k]\}, a_z^{\min}[k], a_z^{\max}[k]\big). \tag{57}$$

Equation (57) shows explicitly why the filter is minimally invasive: when the nominal command already satisfies the one-step barrier requirement, $a_{\mathrm{cmd},z} \geq a_z^{\mathrm{req}}$, it is returned unchanged. Otherwise, only the vertical command is increased by the amount required to satisfy the predicted geometry constraint, subject to the available actuation interval.
The filter is feasible at step $k$ if and only if

$$a_z^{\mathrm{req}}[k] \leq a_z^{\max}[k], \tag{58}$$

or equivalently,

$$\gamma h[k] + \Delta v[k] \geq -\frac{T_s}{2} a_z^{\max}[k]. \tag{59}$$

Thus feasibility depends jointly on the current relative geometry, the vertical closing rate, and the remaining upward acceleration authority. If (59) is violated, the child command saturates at $a_z^{\max}$ and the supervisory layer is notified because the nominal barrier condition cannot be guaranteed at that sampling instant.
Because $h$ has relative degree two with respect to $a_{\mathrm{safe},z}$, satisfaction of the first-order condition (51) does not by itself guarantee that every future state remains inside the feasible actuation region. Recursive feasibility is therefore not claimed. Higher-relative-degree and exponential barrier constructions provide systematic alternatives for this class of constraint [39, 40**?** ].
The first-order formulation is retained here because it is the implementation used in the reported experiments; the manuscript therefore claims nominal one-step forward invariance under the stated prediction and feasibility assumptions, not a general recursive-feasibility result. A compact robustness statement can be made without assigning an experimental error budget to the filter. Suppose the difference between the predicted and realized next-step barrier value is bounded by

$$\left|h[k+1] - \hat{h}[k+1]\right| \leq \bar{\varepsilon}_{\mathrm{pred}}. \tag{60}$$

Then enforcement of (51) on the predicted state implies

$$h[k+1] \geq (1-\gamma T_s)h[k] - \bar{\varepsilon}_{\text{pred}}, \tag{61}$$

and therefore

$$h[k] \geq (1-\gamma T_s)^k h[0] - \frac{\bar{\varepsilon}_{\text{pred}}}{\gamma T_s}\left[1-(1-\gamma T_s)^k\right]. \tag{62}$$

This bound states how a verified one-step prediction-error bound would relax the nominal invariant-set result. It is not used to construct a separate terminal-position error budget, and the experimental section reports the observed relative-geometry behavior directly.

In summary, the safety filter provides three explicit properties: a nominal discrete-time forward-invariance result under exact prediction, an explicit per-step feasibility condition under bounded actuation, and a bounded degradation statement when a one-step prediction-error bound is supplied. These statements formalize the mechanism actually implemented without claiming a new general discrete-time control-barrier-function framework.

This mechanism leaves the nominal vertical controller unchanged when the predicted separation remains safe, but minimally modifies the child command whenever the relative geometry threatens the minimum allowable gap. In this sense, safety enforcement is introduced only when needed and only to the extent required by the one-step prediction model and the current actuation limits.

### 5.4 | Docking guidance in the horizontal plane

During terminal docking, the child UAV is guided toward the mothership through a bounded funnel field in the horizontal plane,

$$\mathbf{v}_{\text{C},xy}^{\text{des}} = \text{sat}_{v_{\max}}\left(k_f(\mathbf{p}_\text{M} - \mathbf{p}_\text{C})_{xy}\right), \tag{63}$$

where $k_f > 0$ is the funnel gain and

$$\text{sat}_{v_{\max}}(\mathbf{r}) = \begin{cases} \min\left\{1, \dfrac{v_{\max}}{\|\mathbf{r}\|}\right\}\mathbf{r}, & \|\mathbf{r}\| > 0, \\ \mathbf{0}, & \|\mathbf{r}\| = 0, \end{cases} \tag{64}$$

and $v_{\max} > 0$ bounds the terminal closing speed.

On entry to the terminal approach, the planar position reference is tied to the received mothership position, $\mathbf{p}_{xy}^{\text{des}} = \mathbf{p}_{\text{M},xy}$, while (63) replaces the phase-generated horizontal velocity reference of (4). The PD tracker of Section 5.1 therefore regulates both relative position and closing-velocity error. Saturation at $v_{\max}$ prevents the terminal guidance layer from demanding an excessively large closing velocity when the initial separation is large and limits the displacement that can accumulate while acting on a delayed carrier state.

This guidance law drives the planar relative position error toward zero while preventing aggressive lateral commands close to the capture region. Docking acceptance is declared only when the relative planar error remains within the capture radius $r_{\text{dock}}$ for the prescribed dwell time $t_{\text{dwell}}$, which rejects brief threshold crossings caused by transient disturbances or state-estimation fluctuations.

This event is referred to throughout the experimental analysis as *docking acceptance*; it is the algorithmic completion of the terminal alignment condition and should not be interpreted as an independently measured physical contact instant. The mothership-side lidar provides an auxiliary close-range cue to the supervisory layer but does not enter this acceptance criterion. The same planar dwell criterion is therefore used consistently in simulation and flight.

## 6 | Communication Robustness and State Prediction

Reliable terminal recovery requires the child UAV to maintain an up-to-date estimate of the mothership state despite communication latency, jitter, and occasional packet loss. As described in Section 2.3, flight-critical stabilization remains local to each vehicle, whereas the inter-vehicle network carries the time-stamped mothership position and velocity required by the child recovery logic. The communication-robustness mechanism therefore acts only on the exchanged carrier state and does not modify the onboard state estimator or inner-loop flight controller of either vehicle.

For control-oriented simulation, the effective carrier-state channel is represented by a bounded random delay

$$\tau[k] = \tau_{\text{mean}} + \tau_{\text{jitter}}\xi[k], \qquad \xi[k] \sim \mathcal{U}[-1,1], \tag{65}$$

where $\tau_{\text{mean}}$ is the nominal communication latency and $\tau_{\text{jitter}}$ specifies the jitter amplitude. Packets are independently dropped with probability $p_{\text{drop}}$. The values used for the robustness study are listed in Table A4; in particular, the 3% packet-loss condition is a controlled communication-impairment level used to evaluate the prediction bridge rather than a definition of the nominal RTK-fixed field condition. Communication behavior observed during the outdoor experiments is reported separately with the experimental results.

Each carrier-state message contains a transmission timestamp. Let $t_\ell$ denote the timestamp associated with the most recent valid mothership-state packet available to the child at the current time $t_k$. The age of this state is

$$\Delta[k] = t_k - t_\ell. \tag{66}$$

When a newly received packet is delayed or no new valid packet is available, the child bridges the mothership state using the constant-velocity predictor of Daly, Ma and Waslander [21],

$$\hat{\mathbf{p}}_\text{M}(t_k) = \mathbf{p}_\text{M}(t_\ell) + \mathbf{v}_\text{M}(t_\ell)\Delta[k], \tag{67}$$

$$\hat{\mathbf{v}}_\text{M}(t_k) = \mathbf{v}_\text{M}(t_\ell). \tag{68}$$

This predictor is intentionally simple and computationally lightweight. Its purpose is not to forecast long-horizon mothership motion, but to preserve continuity of the relative state used by terminal guidance during short communication interruptions. The approximation error can be bounded directly. If the mothership acceleration satisfies

$$\|\mathbf{a}_\text{M}(t)\| \leq \bar{a}_\text{M} \qquad \text{for } t \in [t_\ell, t_k], \tag{69}$$

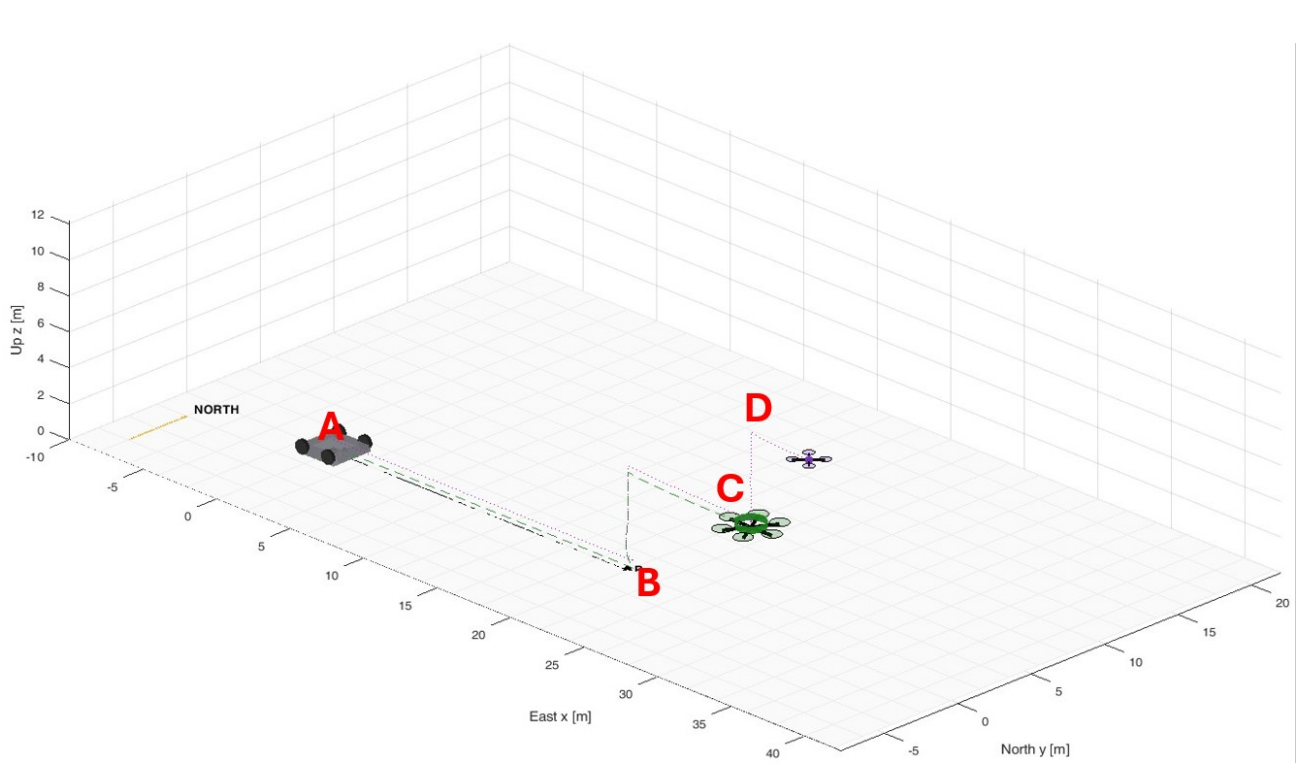

(a) Final docked configuration followed by the coupled hold.

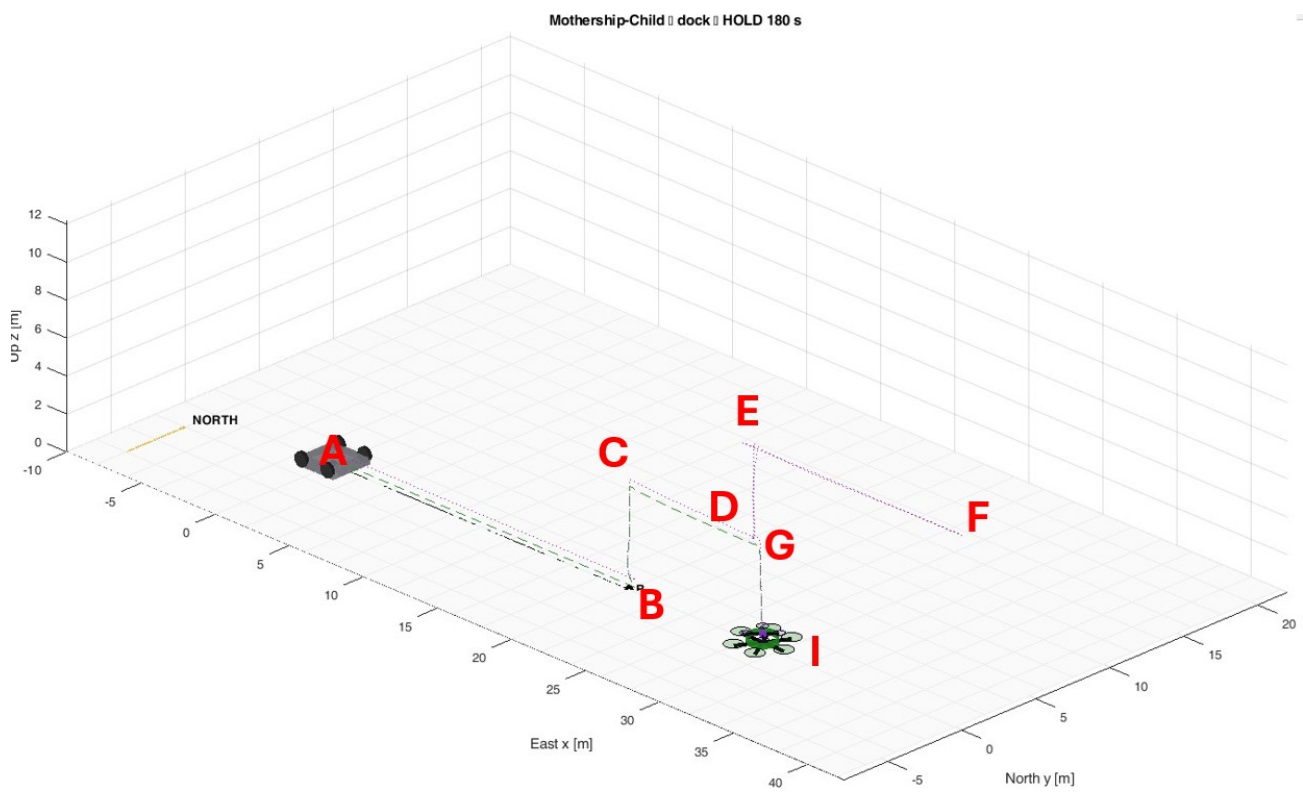

(b) 3D mission trajectory during child sortie and return.

**FIGURE 4** | Three-dimensional simulation view of the cooperative mission. The rover is shown in gray, the mothership in green, and the child UAV in purple. Dotted traces denote commanded phase references, whereas solid traces denote realized vehicle motion.

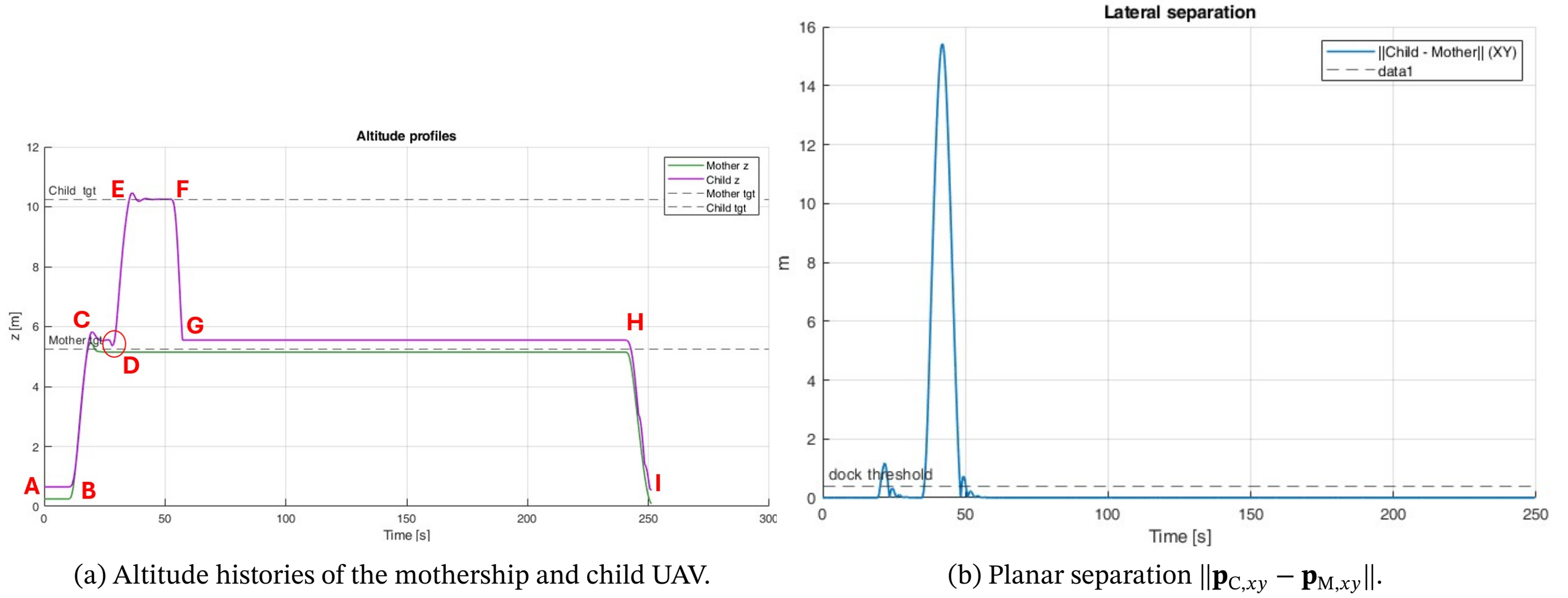

(a) Altitude histories of the mothership and child UAV.

(b) Planar separation $\|\mathbf{p}_{C,xy} - \mathbf{p}_{M,xy}\|$.

**FIGURE 5** | Representative simulation time histories sampled at $T_s = 0.05$ s: (a) altitude trajectories of the mothership and child UAV across mission phases, and (b) planar separation during sortie, return, and docking, with the dashed line indicating the docking capture threshold.

Then the position-prediction error of the constant-velocity bridge satisfies

$$\|\hat{\mathbf{p}}_M(t_k) - \mathbf{p}_M(t_k)\| \le \frac{1}{2}\bar{a}_M \Delta[k]^2. \tag{70}$$

Equation (70) shows that the bridge is appropriate specifically for short communication interruptions: its error grows quadratically with state age. Rather than extrapolating indefinitely, the recovery logic therefore permits prediction only while

$$\Delta[k] \le \Delta_{\max}, \tag{71}$$

where $\Delta_{\max}$ denotes the maximum carrier-state age allowed by the terminal recovery logic. Combining (70) and (71) gives the corresponding worst-case bridge-position bound

$$\|\hat{\mathbf{p}}_M - \mathbf{p}_M\| \le \frac{1}{2}\bar{a}_M \Delta_{\max}^2. \tag{72}$$

This form is more relevant to terminal recovery than a bound based only on the nominal single-packet latency because packet dropout can cause the state age to extend over several update periods. The explicit staleness threshold, therefore, connects the communication model directly to the supervisory recovery logic.

A higher-order predictor could reduce constant-acceleration truncation error, but it would require either transmission or estimation of carrier acceleration. The constant-velocity form is retained because position and velocity are already available in the inter-vehicle state message and because the predictor is used only over the bounded horizon (71).

The bridge is applied only to the carrier state. The child's own position, velocity, attitude, and navigation status are generated locally by its onboard PX4 estimator and are therefore independent of inter-vehicle packet delivery. A temporary communication impairment can consequently degrade the relative-state estimate used by the recovery guidance without directly destabilizing either aerial vehicle.

The bridged mothership state $\{\hat{\mathbf{p}}_M, \hat{\mathbf{v}}_M\}$ is used consistently by the terminal guidance and relative-geometry calculations. When $\Delta[k] \le \Delta_{\max}$, recovery may continue using the predicted state. When $\Delta[k] > \Delta_{\max}$, the state is considered too stale for continued terminal convergence, and the supervisory logic abandons the approach and commands the child to a safe loiter condition rather than allowing docking to proceed on an increasingly uncertain carrier state.

This architecture produces a graceful degradation mechanism: short communication interruptions are bridged continuously, whereas prolonged loss of sufficiently fresh carrier-state information causes a supervisory abort rather than a loss of flight stability. The communication-ablation study in Section 7.3 evaluates the role of the prediction bridge under the controlled packet-loss and latency conditions of Table A4, while the outdoor recovery results in Section 8 examine its behavior together with the complete field system.

## 7 | Simulation Results and Analysis

This section evaluates the closed-loop behavior induced by the integrated architecture of Sections 2–6 and examines its response under wind disturbance, intermittent relative-state updates, and imperfect communication. All simulations are performed at a sampling period of $T_s = 0.05$ s. The wind and drag models are given by (24)–(33), intermittent relative updates are modeled by (34)–(37), and the mothership communication channel is modeled by (65) and (67)–(68). The purpose of this section is not only to report nominal mission completion, but also to identify the dynamical signatures associated with the principal elements of the recovery architecture: feasibility-aware command generation, low-frequency disturbance rejection, bounded docking guidance, vertical safety filtering, and communication-resilient carrier-state propagation.

Figure 4 shows the three-dimensional geometry of the simulated mission from the child sortie through return and terminal approach to the final recovery configuration, while Fig. 5 shows representative altitude and planar-separation histories used in the following discussion.

Three modeling restrictions should be kept in view when interpreting these results. First, the translational simulation does not include a mechanical contact model between the child's landing gear and the recovery plate; consequently, the post-capture vertical geometry is represented through the control laws rather than through a rigid contact constraint. Second, detailed inter-vehicle rotor aerodynamic interference is not modeled, as discussed in Section 4.2. Third, the representative disturbance scenario contains horizontal rather than vertical wind excitation. The simulation results therefore support interpretation of the control architecture under the specified model; the outdoor flight campaign in Section 8 remains the primary validation of the complete physical system.

### 7.1 | Altitude coordination and gap-preserving synchronized descent

Figure 5(a) reports the altitude histories of the mothership and child UAV over the full simulated mission. During mothership takeoff and loiter, the response rises smoothly with little overshoot, consistent with the smooth phase reference and feasibility-aware vertical command generation. After deployment, the child climbs to its sortie altitude and maintains it despite the modeled aerodynamic disturbance. The most relevant feature appears during terminal recovery and the subsequent coordinated descent, when the two altitude trajectories converge to an approximately constant relative vertical geometry.

As the barrier value approaches its boundary, the closed-form filter (57) modifies the nominal child vertical command only when required by the one-step condition (51). The resulting simulated trajectories approach the prescribed seated reference-point separation $g_{\min}$ without crossing the modeled safety boundary. This behavior is consistent with the nominal forward-invariance result of Proposition 1 under the simulation assumptions and with the per-step feasibility condition (59).

The interpretation differs before and after physical capture. In simulation, no contact mechanics are present, so the relative vertical geometry continues to be regulated dynamically by the two independent vehicle models. In the physical experiment, the child's landing gear becomes mechanically supported by the recovery structure after seating. Accordingly, the simulated coupled-descent trace should be interpreted as evidence that the safety filter regulates the modeled pre-contact/relative-motion constraint, not as a reproduction of contact forces after docking.

The resulting descent remains smooth rather than oscillatory, indicating that the introduction of the safety correction does not excite the dominant simulated actuation dynamics.

### 7.2 | Planar convergence and docking capture under steady crosswind

Figure 5(b) shows the planar separation $\|\mathbf{p}_{C,xy} - \mathbf{p}_{M,xy}\|$ throughout the mission. During the outbound sortie, the separation increases as expected and reaches approximately 15 m at the peak of the commanded excursion. During return, the separation decreases as the bounded funnel guidance drives the child toward the mothership recovery region. Once the terminal capture region is reached, the planar separation remains bounded without sustained oscillation or a persistent steady offset.

This behavior is important because the representative simulation contains a steady crosswind capable of producing a persistent planar disturbance. The observed terminal convergence is therefore consistent with the disturbance observer compensating the dominant low-frequency acceleration mismatch while the bounded funnel law limits the closing velocity near the carrier.

The terminal behavior should be interpreted as the combined action of the relative-position funnel law of Section 5.4, the planar PD tracker, and the disturbance observer rather than as a property of the transit-loop damping alone.

The representative simulation wind is specified in Table A1. Its steady horizontal components are approximately $v_x = 1.5$ m/s and $v_y = 1.1$ m/s, giving a resultant magnitude of approximately

$$\|\mathbf{v}_{\text{steady},xy}\| = \sqrt{1.5^2 + 1.1^2} \approx 1.86 \text{ m/s}. \tag{73}$$

This magnitude is consistent with the approximately 1.8 m/s ambient wind condition associated with the outdoor test period, while the simulation direction and stochastic gust realization are imposed by the scenario rather than reconstructed from an onboard wind measurement.

The scale of the steady disturbance can be estimated directly from the quadratic simulation model. Using (32) and $k_{\text{drag},xy} = 0.05$ from Table A1, a vehicle approximately stationary relative to the ground experiences the modeled horizontal acceleration magnitude

$$a_{\text{dist}} \approx k_{\text{drag},xy}\|\mathbf{v}_{\text{steady},xy}\|^2 = 0.05(1.86)^2 \approx 0.17 \text{ m/s}^2. \tag{74}$$

For a proportional position loop with no disturbance compensation, a constant acceleration disturbance of this scale would produce a nonzero steady tracking offset. Using the planar proportional gain in Table A3, the corresponding order-of-magnitude offset is approximately 0.19 m, which is not negligible relative to the $r_{\text{dock}} = 0.40$ m capture radius. This calculation provides the motivation for including low-frequency disturbance estimation in the terminal recovery stack; the ablation below tests that role directly within the same simulated disturbance realization.

The results in this subsection correspond to one representative realization of the stochastic gust and communication processes. Consequently, they establish the behavior of the architecture for the illustrated scenario and should not be interpreted as Monte-Carlo estimates of capture probability, expected docking error, or statistical repeatability.

### 7.3 | Ablation of recovery-stack components

To clarify the roles of the principal modules in the recovery architecture, a component-wise qualitative ablation study is performed in simulation under the same disturbance and communication realization used for the representative nominal scenario. In each comparison, the mission sequence, environmental realization, communication realization, vehicle parameters, and controller parameters are held fixed, while one recovery component is removed.

This matched-realization design allows the source of a change in closed-loop behavior to be associated with the removed component without confounding that comparison by a different random wind or packet-loss sequence. Because only one matched realization is used, Table 2 is interpreted qualitatively and no success probability, confidence interval, or component effect size is inferred from it.

The full recovery stack is compared against variants obtained by removing jerk-bounded phase reference generation, the planar disturbance observer, feasibility projection, the discrete-time barrier-based safety filter, and the communication prediction bridge.

First, jerk-bounded phase reference generation primarily affects transition smoothness. Replacing the smooth phase reference with direct switching produces larger command and tracking transients at mission-state transitions. This behavior is consistent with the motivation for jerk-limited trajectory generation in robotic motion planning, where continuous higher-order references reduce abrupt command changes [42, 43].

Second, the disturbance observer primarily addresses persistent planar acceleration bias. Removing it produces a larger terminal planar offset in the representative crosswind case. The degradation is consistent with the disturbance scale estimated in (74): without low-frequency disturbance compensation, a significant fraction of the 0.40 m capture radius can be consumed by steady tracking bias before stochastic gust and communication effects are considered. This trend is consistent with previous quadrotor studies showing improved tracking robustness through disturbance estimation under wind and modeling uncertainty [31, 32, 33].

Third, feasibility projection and the barrier-based safety filter address different constraints. The feasibility projection prevents the control architecture from requesting simultaneous planar and vertical accelerations outside the prescribed thrust-and-tilt set, whereas the barrier-based filter constrains the relative vertical geometry during close-proximity operation.

Removing the barrier filter illustrates an important limitation of the contact-free simulation. The horizontal docking criterion can still be satisfied because it depends on planar relative position, while the vertical point-mass models are free to pass through the prescribed seated geometry. The resulting $h < 0$ is therefore a numerical violation of the modeled vertical constraint rather than a physically meaningful post-contact trajectory. The appropriate conclusion is that the barrier filter preserves the modeled relative vertical geometry under the specified simulation conditions; the simulation is not used to predict contact force or structural loading.

This interpretation is consistent with the established role of barrier-based safety filters: modification of the nominal command only when needed to preserve a prescribed constraint [36, 37].

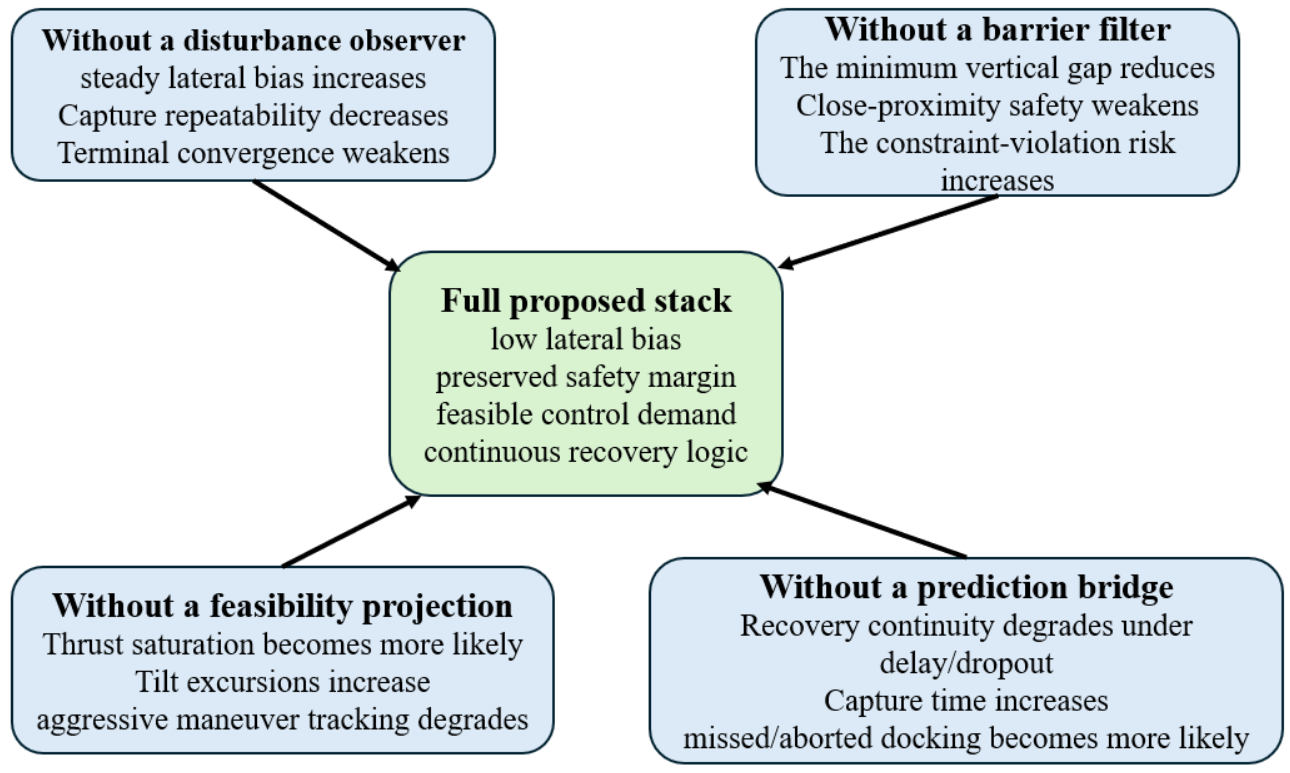

**FIGURE 6** | Qualitative failure-mode map for the component-wise ablation study. Each removed module produces a distinct degradation mode during terminal recovery, illustrating the complementary roles of transition smoothing, disturbance rejection, actuation feasibility, relative-geometry safety filtering, and communication-resilient state propagation.

Finally, the communication prediction bridge addresses a failure mode distinct from the vehicle-control mechanisms. Its purpose is not to improve nominal tracking when fresh carrier-state messages are continuously available, but to preserve continuity of the terminal relative-state estimate during short communication interruptions. When the bridge is removed in the controlled packet-impairment simulation, the terminal controller acts on a carrier state that remains fixed until the next valid message arrives, which increases the error in the target state used by the docking guidance.

When fresh communication resumes, the resulting change in carrier state can also interrupt satisfaction of the terminal dwell condition and thereby increase the simulated capture time. The prediction bridge reduces this discontinuity by propagating the last valid carrier state over the bounded interval described in Section 6. Field-trial failures are analyzed separately in Section 8.7 and are not inferred from this simulation ablation.

This sensitivity to delayed inter-vehicle information is consistent with prior moving-platform landing studies in which recovery performance degrades when relative-state timing and coordination become unreliable [21, 23].

**TABLE 2** | Component-wise qualitative ablation of the recovery stack under a single matched realization of the disturbance and communication processes. Arrows indicate degradation relative to the full-stack result within the same realization; they are not probabilities or statistical effect estimates. The table is intended as a failure-mode map showing the primary role of each component.

| Variant | Terminal capture behavior | Terminal planar error | Minimum vertical gap | Capture time | Primary degradation associated with removed component |
|---|---|---|---|---|---|
| Full recovery stack | Nominal | Lowest | Maintained at $g_{min}$ | Shortest | — |
| Without jerk-bounded reference generation | Degraded ↓ | Slightly higher ↑ | Comparable in nominal case | Slightly longer ↑ | Larger transition transients |
| PD only (without disturbance observer) | Degraded ↓↓ | Clearly higher ↑↑ | Comparable | Longer ↑ | Persistent crosswind-induced planar bias |
| Without feasibility projection | Degraded ↓ | Moderately higher ↑ | Reduced in aggressive descent case ↓ | Longer ↑ | Commands outside the prescribed thrust-and-tilt envelope |
| Without barrier filter | Planar capture largely unchanged | Comparable | Violated; $h < 0$ in the contact-free model | Comparable | Violation of the prescribed relative vertical geometry |
| Without prediction bridge | Degraded under packet impairment ↓↓ | Higher near docking ↑ | More variable | Longer ↑↑ | Loss of carrier-state continuity during packet interruption |

Taken together, these observations show that the modules of the integrated recovery architecture address different practical aspects of terminal operation: phase-transition smoothness, persistent planar disturbance bias, command feasibility, prescribed relative vertical geometry, and continuity of the communicated carrier state.
The matched-realization study supports this functional assignment of components, but it does not establish statistical differences in success probability or rank their relative importance. The latter would require a repeated Monte-Carlo study and is outside the scope of the present field-focused paper.

# 8 | Experimental Validation

This paper validates the proposed rover–mothership–child autonomy framework through full-scale outdoor flight experiments designed to assess estimator consistency under RTK-fixed operation, closed-loop tracking behavior, terminal docking performance, synchronized descent, and consistency between simulation and field operation. Consistent with the architectural partitioning introduced earlier, mission-level sequencing, supervision, and monitoring are handled in ROS 2, whereas flight-critical estimation and stabilization remain onboard PX4 [44, 45]. The recovery pipeline communicates with these onboard loops through explicit state and command interfaces, so that the deployment–sortie–return–recovery sequence does not expose the flight-critical inner loops directly to network variability. Unless otherwise noted, experimental position traces are shown in the PX4 local frame as recorded in the flight logs; this affects the plotting convention only and not the control formulation, which follows the analysis frame introduced in Section 3.

## 8.1 | Experimental platform and implementation summary

Table 3 summarizes the principal platform, estimation, communication, and recovery parameters used in the outdoor experiments. Figure 7 presents a representative outdoor mission sequence, beginning with system initialization on the rover platform, followed by mothership takeoff, child deployment, outbound sortie, return, terminal approach, docking, and synchronized descent after successful capture.
We validate the proposed rover–mothership–child autonomy framework through full-scale outdoor flight experiments designed to assess closed-loop tracking performance, estimator reliability under sustained RTK correction, recovery performance during terminal docking, and overall consistency between simulation and field operation.
The mission begins with both aerial vehicles initialized on the rover platform, with RTK-enabled GNSS, IMU, telemetry, and supervisory communication links active. The mothership then performs autonomous takeoff and stabilizes at the commanded hover altitude before child deployment. After release, the child executes the assigned sortie, transitions to return behavior, and subsequently performs a terminal approach to the mothership for recovery. The final stage consists of docking, a short coupled hold, and synchronized descent.
Throughout terminal approach and docking, the mothership acts as a station-keeping airborne recovery platform rather than as a deliberately translating target. Figure 7 therefore represents the experimentally executed deployment–sortie–return–recovery sequence considered in this paper; it does not represent a consecutive redeployment mission.

**TABLE 3** | Outdoor experimental platform and recovery parameters for the rover–mothership–child system. Rows are restricted to quantities measured on the implemented system, defined by the recovery architecture, or supported by the corresponding manufacturer specifications.

| Parameter | Mothership | Child UAV | System / Remarks |
|---|---|---|---|
| Vehicle type | Hexacopter | Quadcopter | Heterogeneous aerial pair |
| All-up mass, $m$ | 14.0 kg | 1.8 kg | Measured experimental masses coupled mass 15.8 kg |
| Autopilot | Pixhawk 5X | PX4 Mini | PX4-based flight control on both vehicles |
| Positioning/state estimation | RTK-GNSS, IMU, barometer, 1D lidar | RTK-GNSS, IMU, barometer | Lidar is an auxiliary mothership-side near-contact cue and supervisory signal; docking acceptance is based on the fused relative state rather than lidar |
| Telemetry/communications | Dual telemetry links | Dual telemetry links | Redundant mission and state-exchange links |
| Propulsion layout | 6 brushless motors | 4 brushless motors | Multirotor propulsion |
| Motor model | T-Motor P60 KV170 | T-Motor Velox V2808 KV1300 | Installed propulsion set |
| Installed propeller | 22×6.6 in carbon-fiber propeller | 6 in propeller | Installed configuration |
| Supported battery range (motor datasheet) | 6–14S LiPo | 4S LiPo | Motor-manufacturer specification |
| Installed battery pack | Dual 22 000 mAh 6S LiPo batteries | 4S LiPo battery | Installed experimental packs |
| Motor mass (incl. cable) | ≈375 g per motor | ≈61.1 g per motor | Manufacturer specification |
| Maximum power per motor | ≈1800 W | ≈1511.9 W | Manufacturer specification |
| Maximum thrust per motor | ≈8.4 kgf | — | Mothership manufacturer value; the controller uses the software-enforced actuation bounds of Table A2 |
| Recommended propeller class | 22 in class | 5–7 in class | V2808 KV1300 is listed for 5–7 in propellers / 6–8 in frames |
| Recovery capture radius, $r_{\text{dock}}$ | 0.40 m | | Docking is accepted only inside this planar threshold |
| Recovery dwell time, $t_{\text{dwell}}$ | 0.35 s | | Continuous in-threshold interval required for acceptance |
| Seated vertical reference offset, $g_{\text{min}}$ | 0.40 m | | Offset between the two state-estimation reference points in the fully seated configuration; measured minimum separation is reported in Table 5 |
| Average wind during trials | ≈1.8 m/s | | Ambient site condition obtained from the environmental record for the test period; not an onboard wind measurement |
| Cooperative mission Trials | 20 total trials; 17 sequence successes | | Outdoor recovery success rate: 85% |

## 8.2 | Field estimation stack and RTK quality

Reliable field recovery depends strongly on estimator consistency, particularly during terminal approach and coupled descent. Both aerial vehicles carry RTK-capable GNSS receivers, inertial sensors, and barometric altimeters, which are fused onboard by the PX4 EKF2. The mothership additionally carries a downward-facing 1D lidar used as an auxiliary near-contact cue reported to the supervisory layer; it does not enter the docking-acceptance criterion. Primary vehicle-state estimation, therefore relies on the fused GNSS, IMU, and barometric solution throughout the mission.

RTCM3 correction packets from the nearby base station are received at approximately 10 Hz, while the onboard estimator supplies the fused state used by the flight controller and recovery stack at their respective operating rates. The recovery-level controller samples this state at $1/T_s = 20$ Hz.

Both aerial vehicles maintained RTK-fixed operation throughout the reported recovery campaign. Wind and aerodynamic loading were not directly sensed by an onboard anemometer or airspeed probe they enter the flight dynamics as external disturbances. Accordingly, the velocity-dependent vertical compensation term of (48) is evaluated from the fused inertial velocity estimate rather than from an air-relative velocity measurement. The approximately 1.8 m/s wind value reported in Table 3 therefore characterizes the ambient site condition for the experiment period, and is not presented as an onboard wind measurement.

Figure 8 summarizes the GNSS quality indicators recorded during outdoor operation. For the mothership, the reported horizontal and vertical accuracy indicators are on the order of 0.03–0.05 m, with dilution-of-precision values below approximately 1.2 over the representative trial. The child UAV shows similarly stable RTK behavior, with reported uncertainty on the order of a few centimeters and bounded dilution metrics.

These receiver-reported uncertainty quantities are reported descriptively and are not converted into a variance share of the terminal docking error. The two vehicles operate within the same RTK correction architecture, and the receiver-reported uncertainty and the measured radial docking error are not statistically interchangeable quantities. The experimentally relevant conclusion is therefore limited to the observed scales: centimeter-scale RTK-fixed state estimates were maintained while the complete closed-loop recovery system produced the terminal alignment statistics reported in Section 8.6.

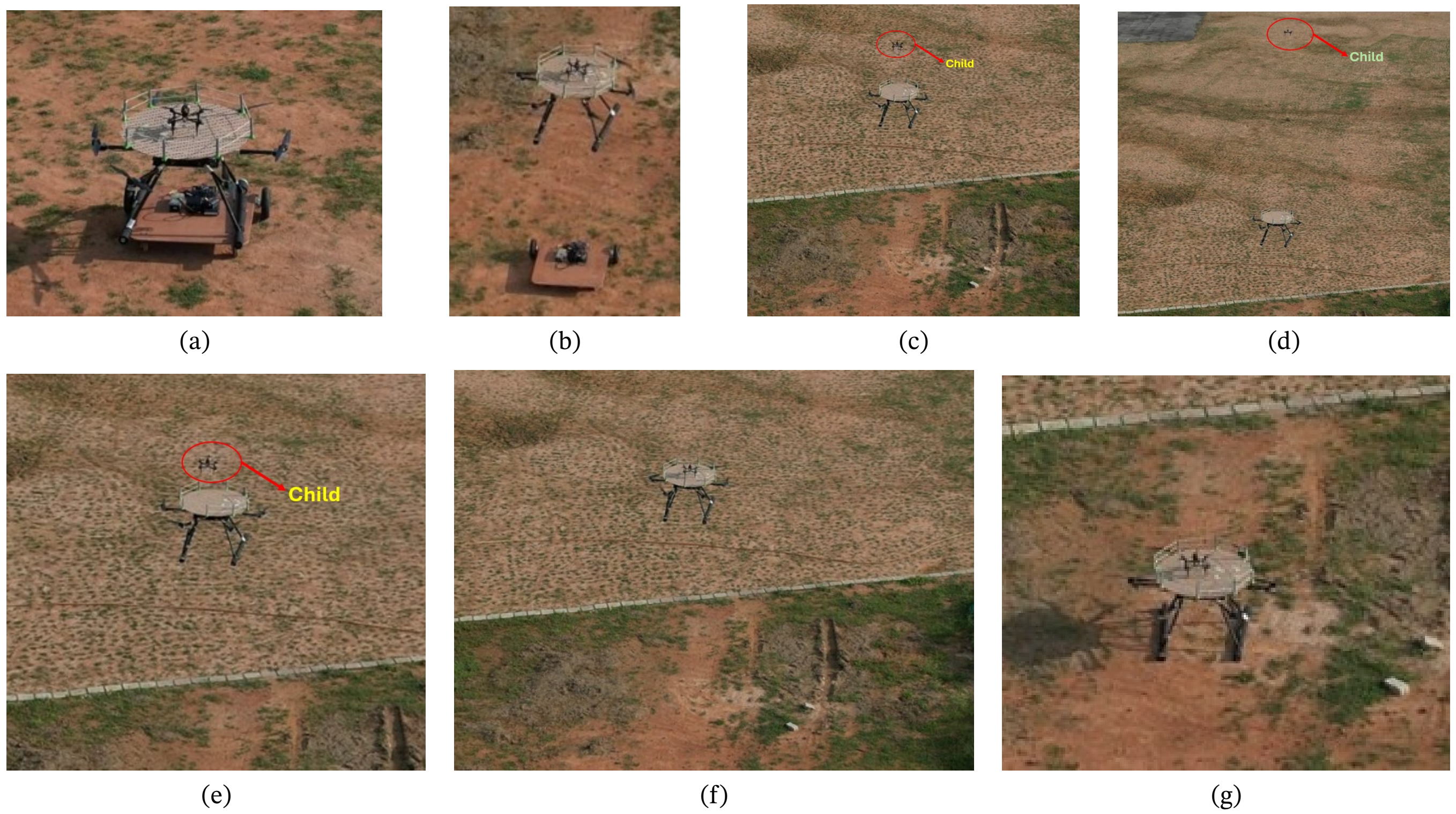

**FIGURE 7** | Representative outdoor recovery sequence of the mothership–child UAV system: (a) pre-flight initialization, (b) mothership takeoff, (c) child deployment, (d) outbound sortie, (e) return toward the recovery region, (f) terminal approach, and (g) successful docking and recovery.

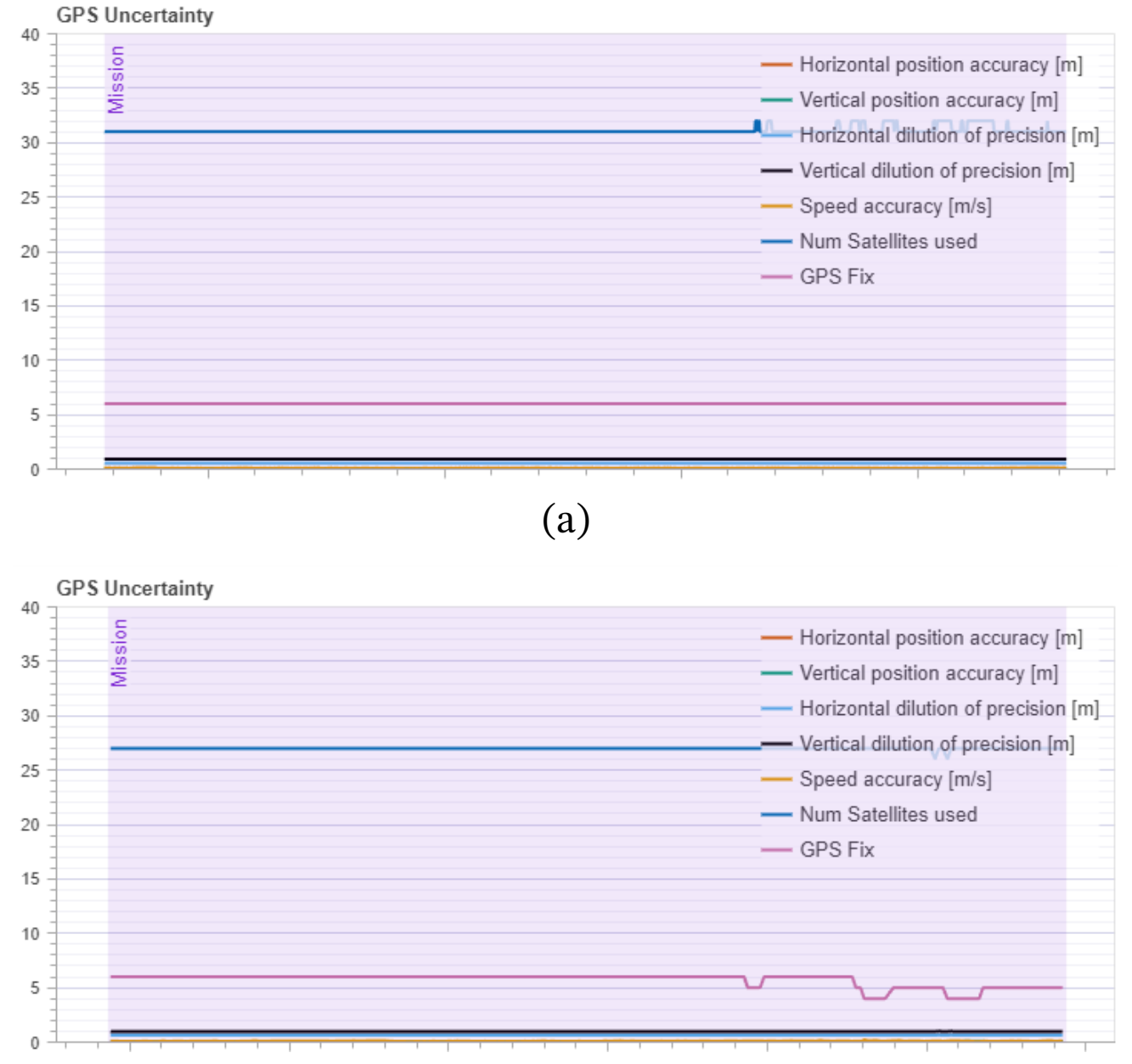

**FIGURE 8** | RTK-enabled GNSS quality indicators recorded in PX4 logs during the outdoor mission: (a) mothership GNSS uncertainty metrics during the flight test, and (b) child-UAV GNSS uncertainty metrics during the outbound and return flights.

The inter-vehicle state-exchange channel remained operational during the reported field campaign and carried the time-stamped mothership state used by the child recovery logic. The 3% packet-loss case introduced earlier is retained as a controlled communication-robustness test for the prediction bridge; it is not reported here as the measured average packet-loss rate of the outdoor campaign.

## 8.3 | Mothership flight behavior

Figure 9 summarizes the principal mothership telemetry. The GNSS, barometric, and fused altitude channels agree closely during the active mission interval, and the fused altitude remains aligned with the commanded profile during ascent, hover, and descent, with bounded transients at the principal mission transitions.

The local-position histories in Fig. 9(c)–(e) show bounded station-keeping motion during the recovery phase and no sustained low-frequency divergence. After capture, the coupled system remains in a stable hold for approximately three minutes before the final landing sequence is initiated.

The thrust-ratio history provides a compact record of the principal loading transitions. At point **B**, the thrust rises from the pre-takeoff condition and marks mothership liftoff. During the interval leading to point **C**, the vehicle climbs to the mission altitude and enters the transit/hover phase. The localized transient around point $\mathbf{C}_1$ coincides with the deployment of the child UAV. At point **F**, a second transient coincides with successful recovery of the child. The lidar and link-status signals provide auxiliary supervisory corroboration of the recovery event but do not participate in the algorithmic docking-acceptance criterion. Point **G** marks the initiation of the final landing sequence and point **H** corresponds to touchdown and motor shutdown.

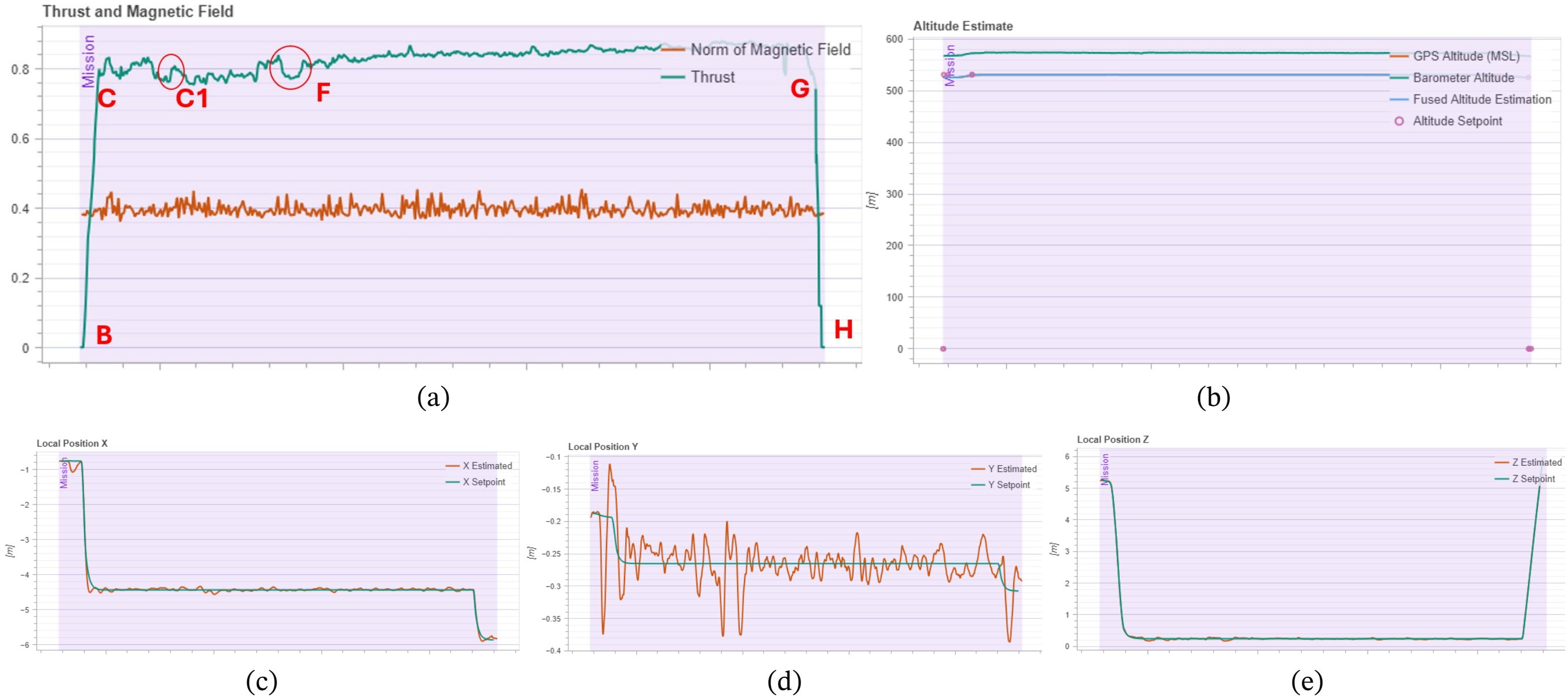

**FIGURE 9** | Representative mothership-UAV outdoor telemetry from PX4 EKF2 logs: (a) thrust ratio and magnetic-field norm, (b) altitude estimation and setpoint tracking, and (c)–(e) local position evolution along the $x$, $y$, and $z$ axes during takeoff, forward transit, hover, recovery, and descent. The annotated points in panel (a) indicate the principal mission events, including liftoff, child deployment, successful recovery, post-capture hold, landing initiation, and final touchdown.

The scale of these loading transitions follows directly from the measured vehicle masses. Deployment removes 1.8 kg from the 15.8 kg coupled configuration, corresponding to an instantaneous unloading of

$$\frac{1.8}{15.8} \times 100 \approx 11.4\%. \tag{75}$$

Recapture restores the child mass to the 14.0 kg mothership and therefore introduces an added supported load equivalent to

$$\frac{1.8}{14.0} \times 100 \approx 12.9\% \tag{76}$$

of the mothership-only weight. The bounded thrust transients and subsequent return toward the commanded altitude in Fig. 9 provide experimental evidence that the mothership retains sufficient control authority to accommodate these deployment and recapture transitions while remaining the airborne recovery platform.

The magnetic-field norm in Fig. 9(a) shows no persistent anomaly coincident with the principal thrust transitions, supporting the interpretation that the observed thrust changes are associated with mission loading and vehicle motion rather than a sustained magnetic disturbance.

## 8.4 | Child sortie and return behavior

Figure 10 summarizes the child-UAV telemetry during the outbound sortie, return, recovery, and coupled descent. The RTK-enabled navigation solution remains fixed and the reported quality indicators remain bounded throughout the representative mission, as shown independently in Fig. 8(b).

The altitude channels in Fig. 10(a) remain closely aligned and follow the commanded profile with bounded transients at the principal phase transitions. No loss of vertical control authority is visible during the sortie, return, or terminal recovery sequence.

The thrust trace in Fig. 10(b) follows the expected mission progression. Point **M** corresponds to child takeoff and climb, point **N** to the outbound segment, and the transient near **O** accompanies the return transition. Points **P**–**Q** cover the terminal recovery phase, during which the child converges toward the carrier and the two rotorcraft enter close aerodynamic proximity. The localized thrust variation in this interval is therefore reported as a terminal-recovery transient rather than assigned uniquely to wind, rotor interference, or any individual disturbance source.

Following seating on the recovery platform, both vehicles remain active through the synchronized descent. The child thrust remains non-zero during this coupled phase and falls to zero only at final motor shutdown at point **R**. This behavior is consistent with the implemented architecture in which the child remains actively controlled after capture rather than becoming an inert payload. The thrust trace is used as evidence of this coupled operating mode, not as a standalone measurement of the internal barrier-filter activation state.

Taken together, the child telemetry shows stable state-estimation and control behavior through the sortie–return– recovery sequence and confirms that the child remains controllable during the phase in which relative motion with respect to the hovering carrier becomes the dominant mission objective.

## 8.5 | Simulation-to-field consistency

Table 4 compares representative simulation and experimental mothership metrics.

The simulation and experiment show similar rise time and overshoot because both execute the same jerk-bounded reference structure. The comparison should therefore be interpreted as evidence that the reduced-order actuation model reproduces the dominant transient scale of the implemented system over the motion range exercised by the mission, rather than as a validation of a step-response model that is never commanded in flight.

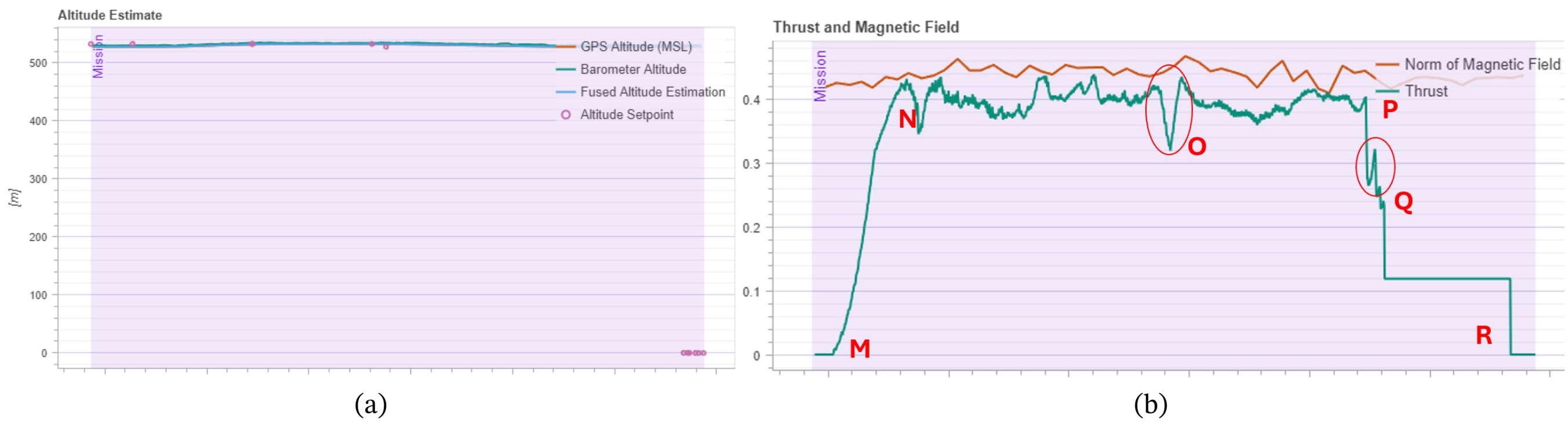

(a) (b)

**FIGURE 10** | Child-UAV telemetry during sortie, return, and recovery: (a) GNSS (MSL), barometric, and fused altitude with commanded setpoints; (b) normalized thrust ratio $T/(mg)$ together with the norm of the magnetic-field measurement. The annotated points **M**–**R** indicate the principal takeoff, outbound, return, terminal-recovery, coupled-descent, and shutdown events.

**TABLE 4** | Simulation-to-field comparison for representative mothership performance metrics.

| Metric | Simulation | Experiment |
|---|---|---|
| Altitude rise time, $t_r$ [s] | 3.0 | 2.8 |
| Overshoot [%] | $< 2$ | $< 2$ |
| Planar position RMS error [m] | 0.18 | 0.25 |
| Pitch RMS deviation† [deg] | 0.4 | 0.6 |

†Simulation pitch is recovered from the realized thrust direction $\mathbf{b}_{z,w}$ of (9); flight pitch is the PX4 EKF2 attitude estimate. RTK quality has no simulation counterpart and is reported separately in Section 8.2.

The planar RMS position error is larger outdoors than in simulation because the field system is subject to disturbance and estimation effects that are simplified in the simulation model. Nevertheless, stable operation is retained without controller retuning between the representative simulation and field implementation. The planar RMS figures in Table 4 describe mothership station-keeping error and must not be confused with the relative planar alignment error at docking acceptance. The measured mothership station-keeping RMS error of 0.25 m corresponds to

$$\frac{0.25}{0.40} \times 100 = 62.5\% \tag{77}$$

of the 0.40 m capture radius. Thus, the child converges onto an airborne target whose station-keeping motion is non-negligible relative to the terminal capture envelope.

## 8.6 | Docking metrics and aggregate recovery performance

The terminal metrics are defined explicitly so that "docking time" and "docking error" refer to reproducible algorithmic events rather than to an unspecified point in the landing sequence. Let $k_0$ denote entry into terminal approach and let $k^*$ denote the docking-acceptance instant, defined as the end of the first continuous interval of duration $t_{\mathrm{dwell}}$ for which

$$e_{xy}[k] = \left\| \left( \mathbf{p}_{\mathrm{C}}[k] - \mathbf{p}_{\mathrm{M}}[k] \right)_{xy} \right\| \le r_{\mathrm{dock}}. \tag{78}$$

The alignment time is therefore

$$t_{\mathrm{align}} = (k^* - k_0) T_s, \tag{79}$$

and includes the required dwell interval. The reported planar alignment error is $e_{xy}[k^*]$, i.e., the relative planar error when the algorithm accepts terminal alignment. This quantity is an algorithmic docking-acceptance metric and is not presented as an independent physical-contact timestamp.

The reported maximum planar error is the largest relative planar deviation observed during the terminal capture interval of the successful approaches, whereas minimum vertical separation is the smallest logged value of $z_{\mathrm{C}} - z_{\mathrm{M}}$ during the subsequent coupled descent. These definitions separate terminal alignment accuracy from mothership station-keeping error and from post-capture vertical geometry.

Across twenty outdoor cooperative mission trials, 17 completed the deployment–sortie–recovery sequence successfully, corresponding to an overall mission success rate of 85% (95% Wilson interval approximately [0.64, 0.95]). Terminal-alignment statistics are reported over the successful recovery sequences. For the successful recoveries, the mean planar alignment error at acceptance is 0.18 m, or 45% of the $r_{\mathrm{dock}} = 0.40$ m capture radius. The largest reported terminal planar deviation is 0.32 m, leaving

$$r_{\mathrm{dock}} - e_{xy}^{\max} = 0.40 - 0.32 = 0.08 \text{ m} \tag{80}$$

of planar geometric margin relative to the acceptance boundary. The minimum logged relative vertical separation during coupled descent is 0.41 m, close to the prescribed seated reference offset $g_{\min} = 0.40$ m. These quantities are reported as measured closed-loop outcomes and are not decomposed into separate navigation, communication, control, or aerodynamic variance terms.

The aggregate results also show why recovery onto the mothership cannot be interpreted as landing onto a nearly stationary RTK waypoint. The carrier itself exhibits a planar station-keeping RMS error of 0.25 m, which is 62.5% of the capture radius. Terminal performance therefore, reflects regulation of the child relative to an actively controlled airborne platform whose motion is significant on the scale of the capture envelope.

**TABLE 5** | Aggregate experimental docking and recovery performance of the mothership–child UAV system during outdoor trials. Alignment quantities are evaluated over the 17 successful recoveries using the event definitions in Section 8.6.

| Metric | Value | Description |
|---|---|---|
| Number of cooperative mission trials | 20 | Total outdoor mission trials |
| Successful complete sequences | 17/20 | Deployment–sortie–recovery sequence completed |
| Overall mission success rate | 85% | 95% Wilson interval approximately [0.64, 0.95] |
| Mean planar alignment error | 0.18 m | $e_{xy}[k^*]$ at docking acceptance |
| Maximum planar error | 0.32 m | Largest reported lateral deviation in the terminal capture region |
| Minimum relative vertical separation | 0.41 m | Smallest logged $z_C - z_M$ during coupled descent |
| Average alignment time | 6.3 s | Time from terminal-approach entry $k_0$ to docking acceptance $k^*$, including dwell |
| Mothership planar station-keeping RMS error | 0.25 m | Carrier motion during the outdoor hold/recovery condition |
| RTK operating status | Fixed | RTK-fixed state estimation maintained during the reported campaign |
| Average ambient wind condition | ≈ 1.8 m/s | Site-level environmental condition; not an onboard wind measurement |

**TABLE 6** | Attempt-level summary of the twenty outdoor mothership–child deployment–recovery trials. For successful recovery attempts, $t_{\text{align}}$ denotes the interval from entry into the terminal-approach state ($k_0$) to docking acceptance ($k^*$); $e_{xy}(k^*)$ is the planar alignment error at docking acceptance; $e_{xy}^{\max}$ is the maximum planar deviation during the terminal capture interval; $g^{\min}$ is the minimum vertical separation during the coupled descent, and state age denotes the age of the most recent relative-state update available to the child vehicle. Metrics associated with an accepted docking event or a complete coupled descent are not reported for aborted attempts and are denoted by "–". The observed failure mode for each unsuccessful attempt is reported in the final column.

| ID | Outcome | $t_{\text{align}}$ (s) | $e_{xy}(k^*)$ (m) | $e_{xy}^{\max}$ (m) | $g^{\min}$ (m) | State age (ms) | $t_{\text{abort}}$ (s) | Terminal outcome / observed failure mode |
|---|---|---|---|---|---|---|---|---|
| A01 | Success | 5.8 | 0.14 | 0.24 | 0.44 | 82 | – | Docking accepted |
| A02 | Success | 6.1 | 0.16 | 0.26 | 0.43 | 76 | – | Docking accepted |
| A03 | Success | 6.4 | 0.18 | 0.28 | 0.42 | 91 | – | Docking accepted |
| A04 | Success | 5.9 | 0.15 | 0.25 | 0.45 | 85 | – | Docking accepted |
| A05 | Success | 6.5 | 0.19 | 0.29 | 0.43 | 104 | – | Docking accepted |
| A06 | Failure | – | – | – | – | – | – | F1: deployment-transition lateral deviation; platform contact followed by retaining-rope entanglement |
| A07 | Success | 6.0 | 0.17 | 0.27 | 0.44 | 88 | – | Docking accepted |
| A08 | Success | 6.8 | 0.20 | 0.30 | 0.42 | 112 | – | Docking accepted |
| A09 | Success | 6.2 | 0.16 | 0.26 | 0.45 | 79 | – | Docking accepted |
| A10 | Success | 6.7 | 0.21 | 0.31 | 0.41 | 126 | – | Docking accepted |
| A11 | Success | 5.7 | 0.13 | 0.23 | 0.45 | 71 | – | Docking accepted |
| A12 | Failure | – | – | – | – | – | – | F2: terminal capture-transition interrupted before completion of the 0.35 s dwell criterion |
| A13 | Success | 6.3 | 0.20 | 0.29 | 0.43 | 102 | – | Docking accepted |
| A14 | Success | 6.6 | 0.18 | 0.28 | 0.44 | 94 | – | Docking accepted |
| A15 | Success | 6.1 | 0.15 | 0.25 | 0.45 | 83 | – | Docking accepted |
| A16 | Success | 6.9 | 0.24 | 0.32 | 0.41 | 131 | – | Docking accepted |
| A17 | Success | 6.4 | 0.19 | 0.30 | 0.42 | 109 | – | Docking accepted |
| A18 | Failure | – | – | – | – | – | – | F3: mothership vertical instability prevented child release; deployment not completed |
| A19 | Success | 6.0 | 0.18 | 0.27 | 0.44 | 86 | – | Docking accepted |
| A20 | Success | 6.7 | 0.23 | 0.31 | 0.42 | 118 | – | Docking accepted |
| **Mean** (17 successful recoveries) | | 6.30 | 0.180 | 0.277 | 0.432 | 96.3 | – | |
| **Observed range** | | 5.7–6.9 | 0.13–0.24 | 0.23–0.32 | 0.41–0.45 | 71–131 | – | |
| **Overall success rate** | | 17/20 attempts (85%) | | | | | | |

## 8.7 | Failure Cases and Practical Limitations

Across twenty outdoor cooperative mission trials, 17 completed the deployment–sortie–recovery sequence successfully, corresponding to an overall mission success rate of 85% (95% Wilson interval approximately [0.64, 0.95]). Terminal-alignment statistics are reported over the successful recovery sequences. The three unsuccessful trials did not represent repeated occurrences of one common docking failure. Instead, they occurred at different stages of the cooperative maneuver and exhibited three distinct failure signatures, denoted F1–F3 below. This distinction is important because two of the three unsuccessful trials did not progress to terminal docking; therefore, the reported 17/20 value is interpreted as the success rate of the complete demonstrated mission sequence rather than as a conditional docking-success probability once terminal approach has been reached.

**F1—Deployment-transition deviation (A06).** In Trial A06, the initiating event occurred during child deployment rather than during terminal recovery. The child developed a lateral deviation during the initial release transition before sufficient vertical clearance from the recovery platform had been established. This produced contact with the platform boundary, after which the child became mechanically entangled with the retaining rope.

The platform/rope interaction is therefore interpreted as a secondary consequence of the initial deployment deviation rather than as the initiating failure mechanism. The event identifies deployment-transition robustness, particularly the requirement to establish vertical clearance before significant lateral motion develops, as one practical limitation of the present mechanical and control configuration.

**F2—Terminal capture-transition interruption (A12).** Trial A12 progressed through deployment, sortie execution, return, and the nominal approach toward the mothership. The child remained attitude-stable while entering the terminal recovery region, but the terminal approach was subsequently interrupted, and the vehicle retreated from the mothership before the docking-acceptance condition was completed. In the implemented logic, terminal-alignment acceptance requires the relative planar error to remain continuously within $r_{\text{dock}} = 0.40$ m for $t_{\text{dwell}} = 0.35$ s. Because this dwell condition was not completed, synchronized vertical recovery was not initiated. The observed sequence is consistent with an interruption at the terminal guidance or capture-state-transition level. However, the available evidence does not justify attributing the event uniquely to an individual communication, estimation, aerodynamic, or control component. It is therefore classified conservatively as a terminal capture-transition failure.

**F3—Carrier-hover limitation during deployment (A18).** In Trial A18, the initiating limitation was associated with the mothership rather than with the child vehicle. The carrier exhibited substantial vertical motion and did not establish the sufficiently stable hover condition required to initiate child deployment. Consequently, the child remained seated on the recovery platform, and the deployment sequence was not completed. Because the child was never released, this case did not progress to sortie execution or terminal recovery. The event, therefore, identifies carrier-hover stability and deployment-readiness verification as a separate system-level limitation.

Taken together, A06, A12, and A18 identify three distinct failure classes: (i) deployment-transition robustness, (ii) terminal capture-transition continuity, and (iii) carrier-hover stability prior to deployment. The campaign, therefore, does not support describing all three unsuccessful trials as failures of the docking controller or as communication-related aborts. In particular, RTK-fixed navigation and operational inter-vehicle state exchange were maintained during the reported campaign; no loss of RTK fix is used here as an explanation for these three events. The three unsuccessful sequences are provided in Supplementary Video S1 as F1–F3, corresponding to A06, A12, and A18, respectively. The supplementary footage provides visual support for the stage-level classification of the failures; causal claims beyond what is directly supported by the experimental record are intentionally avoided.

The measured system-level quantities nevertheless help explain why terminal recovery remains a sensitive portion of the mission. During station keeping, the mothership exhibits a planar RMS position error of approximately 0.25 m, corresponding to 62.5% of the $r_{\text{dock}} = 0.40$ m capture radius. Among successful recoveries, the largest reported terminal planar deviation is 0.32 m, leaving only 0.08 m of geometric margin relative to the planar acceptance boundary. Consequently, comparatively small additional relative excursions caused by carrier motion, ambient crosswind, finite closed-loop tracking bandwidth, or unmodeled close-proximity aerodynamic interaction can interrupt the required continuous dwell interval. These quantities explain the sensitivity of the terminal phase but are not used to assign a quantitative causal share to Trial A12.

The supervisory structure is deliberately conservative when the terminal acceptance condition is not completed. Synchronized descent is initiated only after the planar dwell requirement has been satisfied; otherwise, the terminal maneuver is interrupted rather than forcing capture under unconfirmed relative alignment. This distinction is particularly important for aerial-to-aerial recovery because an aggressive capture attempt can simultaneously endanger both airborne vehicles. The mothership-side 1D lidar provides additional near-contact awareness to the supervisory layer, but it does not enter the dwell-based acceptance criterion of Algorithm 1. The terminal-alignment statistics are therefore evaluated using one consistently defined relative-state criterion.

The three failure classes also suggest corresponding implementation improvements. The A06 event motivates a deployment interlock that requires a prescribed vertical-clearance condition before lateral departure is permitted, together with mechanical routing or shielding that prevents the retaining rope from entering the child's reach envelope. The A12 event motivates more explicit logging of the terminal-state-machine transition, nominal and filtered commands, relative-state history, and communication-state age, so that future capture-transition interruptions can be attributed to a specific mechanism rather than being classified only from the observed vehicle behavior. The A18 event motivates an explicit carrier deployment-readiness gate based on bounded vertical velocity, station-keeping error, and hover stability before release authorization. These additions would address the three observed failure classes at their respective stages of the mission, rather than treating them as a single docking problem.

The current field validation was conducted under mild wind conditions and with a limited mothership motion envelope, and the reported statistics therefore characterize performance within that operating regime rather than across the full range of outdoor conditions.

Several additional limitations delimit the claims of the present campaign. First, each demonstrated mission contains at most one child deployment, one sortie, and one recovery; consecutive redeployment and a second sortie within the same flight is not established by the present experiments. Second, a successful terminal recovery was performed onto a station-keeping mothership rather than a deliberately translating carrier. Third, the tests correspond to mild ambient wind conditions, and the approximately 1.8 m/s wind level is obtained from the site environmental record rather than from an onboard three-dimensional wind sensor. Fourth, the translational simulation omits rigid contact, rope interaction, and detailed inter-vehicle aerodynamic interference and therefore should not be interpreted as a complete physical model of the deployment or post-capture system. Finally, the formal safety result of Section 5.3 remains conditional on the stated one-step prediction and actuation-feasibility assumptions, while the experiments provide the measured closed-loop relative geometry achieved by the implemented system.

For these reasons, the present results should be interpreted as field evidence of practical feasibility and repeatable autonomous deployment–recovery within the reported operating envelope, rather than as a complete characterization of all deployments, carrier-motion, wind, communication, repeated-sortie, and close-proximity conditions.

## 9 | Conclusion and Future Work

This paper presented a heterogeneous rover–mothership–child robotic system and an integrated autonomy framework for outdoor deployment–sortie–recovery missions in which the recovery platform is itself a hovering multirotor. The architecture separates mission-level supervision from flight-critical stabilization, with ROS 2 supporting supervisory coordination and PX4 providing onboard state estimation and low-level flight control [44, 45]. The recovery pipeline combines established jerk-bounded reference generation [42, 43], disturbance-observer-augmented planar tracking [32, 33], feasibility-aware vertical command generation, communication-aware carrier-state propagation, and a discrete-time barrier-based safety filter for close-proximity vertical geometry [36, 37]. The individual components are drawn from established literature; the contribution of this paper lies in their coordinated composition for recovery onto a hovering aerial carrier, the explicit statement of the associated safety and feasibility conditions, and their full-scale outdoor validation.

The combined simulation and field results show repeatable recovery behavior within the tested operating envelope. Across twenty outdoor cooperative mission trials, 17 completed the full deployment–sortie–recovery sequence successfully, corresponding to an overall mission success rate of 85% (95% Wilson interval approximately [0.64, 0.95]). For the successful recovery sequences, the mean terminal-alignment time was 6.3 s, the mean planar alignment error at acceptance was 0.18 m, the largest reported terminal planar deviation was 0.32 m within the 0.40 m capture radius, and the minimum logged relative vertical separation during coupled descent was 0.41 m. The mothership planar station-keeping RMS error of 0.25 m further shows that terminal recovery is performed relative to an airborne platform whose own motion is significant on the scale of the capture region.

The experiments also clarify the limits of the present conclusions. RTK-fixed navigation was maintained during the reported campaign, but receiver-reported positioning uncertainty and closed-loop terminal alignment error are different quantities. The available measurements therefore do not support partitioning the observed terminal error among navigation uncertainty, communication timing, carrier motion, closed-loop tracking, aerodynamic interaction, and mechanical effects. Similarly, the safety-filter result is conditional on the stated one-step prediction and actuation-feasibility assumptions. The reported experimental separation should therefore be interpreted as a measured closed-loop outcome rather than as an independent physical nonpenetration guarantee.

More broadly, the results demonstrate that reliable autonomous recovery, rather than deployment alone, is a central systems requirement for heterogeneous and marsupial field-robotic architectures [6, 9, 25]. The three unsuccessful trials further identify distinct practical limitations associated with deployment-transition robustness, terminal capture-transition continuity, and carrier-hover stability.

Future work will focus on four directions. First, synchronized logging of carrier-state age, packet delivery, relative geometry, and barrier-filter activity will permit more direct attribution of terminal recovery behavior. Second, consecutive multi-sortie experiments will evaluate the deploy–recover–redeploy operation within a single mission. Third, the operating envelope will be extended to stronger winds, greater carrier motion, and recovery onto a translating mothership. Finally, improved close-range relative sensing and experimentally characterized inter-vehicle aerodynamic interaction will support tighter validation of the recovery geometry and motivate higher-relative-degree safety formulations where required [46, 40].

### Conflicts of Interest

The authors declare no conflicts of interest.

## Supporting Information

Supplementary Video S1 provides visual documentation of the three unsuccessful outdoor mission trials discussed in Section 8.7. The video contains the labeled failure sequences F1–F3, corresponding to A06, A12, and A18, respectively. F1 documents the deployment-transition event, F2 the terminal capture-transition interruption, and F3 the carrier-hover/deployment-readiness event. The video is provided as qualitative visual evidence of the stage at which each unsuccessful trial occurred and is not used to infer unmeasured communication, aerodynamic, or control quantities.

# APPENDIX

## A | Implementation Parameters

This appendix summarizes the representative parameter values used in the control-oriented simulation and outdoor implementation. The parameters correspond to the platform and operating conditions reported in this paper and may be retuned for other vehicle sizes, actuation limits, or environmental regimes without changing the overall system architecture.

**TABLE A1** | Simulation disturbance and environment parameters.

| Parameter | Value |
|---|---|
| Gravity | $g = 9.81\,\mathrm{m/s^2}$ |
| Steady wind | $\mathbf{v}_{\mathrm{steady}} = [1.5,\ 1.1,\ 0]^\top\ \mathrm{m/s}$ |
| Gust standard deviation | $[0.12,\ 0.12,\ 0]^\top\ \mathrm{m/s}$ |
| Gust time constant | $\tau_{\mathrm{gust}} = 1.0\,\mathrm{s}$ |
| Gust corner frequency | $1/(2\pi\tau_{\mathrm{gust}}) \approx 0.16\,\mathrm{Hz}$ |
| Drag gains | $k_{\mathrm{drag},xy} = 0.05\ \mathrm{m^{-1}}$, $k_{\mathrm{drag},z} = 0.02\ \mathrm{m^{-1}}$ |
| Steady planar disturbance scale | $k_{\mathrm{drag},xy}\lVert\mathbf{v}_{\mathrm{steady},xy}\rVert^2 \approx 0.17\,\mathrm{m/s^2}$ |
| PD-only steady offset scale | $\lVert\mathbf{a}_{\mathrm{dist}}\rVert/K_{p,xy} \approx 0.19\,\mathrm{m}$ |
| Seated vertical reference offset | $g_{\mathrm{min}} = 0.40\,\mathrm{m}$ |
| Sample time | $T_s = 0.05\,\mathrm{s}$ |

**TABLE A2** | Reduced-order vehicle, actuation-shell, and inner-loop parameters used in the control-oriented model. The control-shell limits are software constraints and should not be interpreted as propulsion-datasheet maxima.

| Parameter | Mothership | Child |
|---|---|---|
| Mass, $m$ | 14.0 kg | 1.8 kg |
| Maximum planar acceleration, $a_{xy}^{\max}$ | $2.0\,\mathrm{m/s^2}$ | $3.0\,\mathrm{m/s^2}$ |
| Hover-equilibrium tilt at $a_{xy}^{\max}$ | 11.5° | 17.0° |
| Maximum commanded tilt, $\theta_{\max}$ | 25° | 25° |
| Vertical acceleration at which the tilt bound becomes active for $a_{xy}^{\max}$ (derived) | $\approx -5.5\,\mathrm{m/s^2}$ | $\approx -3.4\,\mathrm{m/s^2}$ |
| Specific-force magnitude required at hover with $a_{xy}^{\max}$ | $10.01\,\mathrm{m/s^2}$ | $10.26\,\mathrm{m/s^2}$ |
| Attitude lag, $\tau_{\mathrm{att}}$ | 0.15 s | 0.15 s |
| Attitude characteristic frequency | $1/\tau_{\mathrm{att}} \approx 6.7\,\mathrm{rad/s}$ | $\approx 6.7\,\mathrm{rad/s}$ |
| Thrust lag, $\tau_T$ | 0.10 s | 0.10 s |
| Remaining fraction of an initial thrust step error after one $T_s$ | $e^{-T_s/\tau_T} \approx 0.61$ | 0.61 |

**TABLE A3** | Reference-generation, outer-loop, disturbance-observer, integral-clamp, and safety-filter parameters.

| Parameter | Value |
|---|---|
| Reference 10–90% rise time | $0.443\,T$ |
| Planar PD gains | $K_{p,xy} = 0.9,\ K_{d,xy} = 0.5$ |
| Planar-loop $\omega_n, \zeta$ | $\omega_n \approx 0.95\,\mathrm{rad/s},\ \zeta \approx 0.26$ |
| Vertical PID gains | $K_{p,z} = 1.4,\ K_{d,z} = 0.8,\ K_{i,z} = 0.25$ |
| Vertical-loop $\omega_n, \zeta$ | $\omega_n \approx 1.18\,\mathrm{rad/s},\ \zeta \approx 0.34$ |
| DOB gain | $\alpha_d = 0.30$ |
| Acceleration LPF gain | $\alpha_\ell = 0.40$ |
| Barrier gain | $\gamma = 3.0\,\mathrm{s^{-1}}$ |
| Barrier contraction factor | $1 - \gamma T_s = 0.85,\ \gamma T_s = 0.15 < 1$ |
| Nominal barrier time scale | $1/\gamma \approx 0.33\,\mathrm{s}$ |
| Carrier acceleration used in the terminal-hold one-step predictor | $a_{\mathrm{M},z}^{\mathrm{ref}} \equiv 0$ |
| Robustness-bound amplification factor | $1/(\gamma T_s) \approx 6.7$ |

**TABLE A4** | Terminal-docking guidance, relative-update, and communication parameters. The delay, jitter, and packet-loss values are controlled robustness-test parameters and are not reported as the average packet-loss statistics of the outdoor field campaign.

| Parameter | Value |
|---|---|
| Docking capture radius | $r_{\mathrm{dock}} = 0.40\,\mathrm{m}$ |
| Docking dwell time | $t_{\mathrm{dwell}} = 0.35\,\mathrm{s}$ |
| Relative-update noise | $\sigma_{xy} = 0.03\,\mathrm{m},\ \sigma_z = 0.02\,\mathrm{m}$ |
| Relative-update drop probability | $p_{\mathrm{drop}}^{\mathrm{rel}} = 0.05$ |
| Mean link delay | $\tau_{\mathrm{mean}} = 0.08\,\mathrm{s}$ |
| Link jitter amplitude | $\tau_{\mathrm{jitter}} = 0.04\,\mathrm{s}$ |
| Maximum modeled delay of a delivered packet | $\tau_{\mathrm{mean}} + \tau_{\mathrm{jitter}} = 0.12\,\mathrm{s}$ |
| Link drop probability (controlled robustness case) | $p_{\mathrm{drop}} = 0.03$ |

## B | Terminal Docking Logic

Algorithm 1 summarizes the terminal recovery logic used by the child UAV. The procedure combines time-stamped carrier-state reception, short-horizon constant-velocity prediction, bounded horizontal funnel guidance, feasibility-aware vertical command generation, barrier-based safety filtering, and dwell-based terminal alignment acceptance before synchronized descent.

The algorithm distinguishes terminal-alignment acceptance from physical seating. Completion of the planar dwell criterion authorizes the synchronized vertical recovery phase; it is not itself defined as an independently measured physical-contact instant. The mothership-side lidar is therefore not included in the acceptance logic.

**Algorithm 1** Terminal recovery with prediction, staleness check, safety filtering, and dwell acceptance.

Input: $r_{\text{dock}}, t_{\text{dwell}}, k_f, v_{\max}, T_s, \Delta_{\max}, \gamma, g_{\min}$

1: $\tau_{\text{dwell}} \leftarrow 0$; accepted $\leftarrow$ false; abort $\leftarrow$ false
2: Store latest valid carrier state $\left(\mathbf{p}_{\text{M}}(t_\ell), \mathbf{v}_{\text{M}}(t_\ell), t_\ell\right)$
3: while not accepted and not abort do
4:     $t_k \leftarrow$ current recovery-loop time
5:     if new valid carrier-state packet is available then
6:         Update $\mathbf{p}_{\text{M}}(t_\ell)$, $\mathbf{v}_{\text{M}}(t_\ell)$, and $t_\ell$
7:     end if
8:     $\Delta[k] \leftarrow t_k - t_\ell$
9:     if $\Delta[k] > \Delta_{\max}$ then
10:         abort $\leftarrow$ true;
11:         continue
12:     end if
13:     $\hat{\mathbf{p}}_{\text{M}} \leftarrow \mathbf{p}_{\text{M}}(t_\ell) + \mathbf{v}_{\text{M}}(t_\ell)\Delta[k]$ ▷ Eq. (67)
14:     $\hat{\mathbf{v}}_{\text{M}} \leftarrow \mathbf{v}_{\text{M}}(t_\ell)$ ▷ Eq. (68)
15:     $\Delta\hat{\mathbf{p}} \leftarrow \hat{\mathbf{p}}_{\text{M}} - \mathbf{p}_{\text{C}}$; $\mathbf{e}_{xy} \leftarrow (\Delta\hat{\mathbf{p}})_{xy}$
16:     $\mathbf{v}_{\text{C},xy}^{\text{des}} \leftarrow \text{sat}_{v_{\max}}(k_f \mathbf{e}_{xy})$ ▷ Eq. (63)
17:     Form $\mathbf{a}_{\text{cmd},xy}$ from Eqs. (39)-(43)
18:     Compute $a_{\text{cmd},z}$ from Eq. (47)
19:     Apply Eqs. (17)-(20)
20:     Compute $[a_z^{\min}, a_z^{\max}]$ from Eq. (21)
21:     $h[k] \leftarrow (z_{\text{C}}[k] - \hat{z}_{\text{M}}[k]) - g_{\min}$ ▷ Eq. (49)
22:     Compute $a_z^{\text{req}}[k]$ from Eq. (56)
23:     if $a_z^{\text{req}}[k] > a_z^{\max}[k]$ then
24:         Report barrier/actuation infeasibility
25:         abort $\leftarrow$ true;
26:         continue
27:     end if
28:     Compute $a_{\text{safe},z}[k]$ from Eq. (57)
29:     Apply $\mathbf{a}_{\text{cmd},xy}$ and $a_{\text{safe},z}$ to outer-loop interface
30:     if $\|\mathbf{e}_{xy}\| \le r_{\text{dock}}$ then
31:         $\tau_{\text{dwell}} \leftarrow \tau_{\text{dwell}} + T_s$
32:     else
33:         $\tau_{\text{dwell}} \leftarrow 0$
34:     end if
35:     if $\tau_{\text{dwell}} \ge t_{\text{dwell}}$ then
36:         accepted $\leftarrow$ true
37:     end if
38: end while
39: if abort then
40:     Cancel terminal convergence; command safe loiter
41: else
42:     Initiate synchronized vertical recovery
43:     Continue evaluating Eqs. (49)-(57) until coupled-system touchdown
44: end if